\documentclass{article}

\usepackage[final, nonatbib]{neurips_2022}

\pdfoutput=1
\usepackage[utf8]{inputenc} 
\usepackage[T1]{fontenc}    
\usepackage{hyperref}       
\usepackage{url}            
\usepackage{booktabs}       
\usepackage{amsfonts}       
\usepackage{nicefrac}       
\usepackage{microtype}      
\usepackage{amsmath,amsfonts,amsthm,amssymb,float,graphicx,subfigure,listings}
\usepackage{algorithm, algorithmic}
\usepackage{wrapfig}
\usepackage{xcolor}         

\newcommand{\mcal}[1]{\mathcal{#1}}

\newcommand{\bsyb}[1]{\boldsymbol{#1}}

\title{Provable General Function Class Representation Learning in Multitask Bandits and MDPs}

\author{%
  Rui Lu$^1$, Andrew Zhao$^1$, Simon S. Du$^2$, Gao Huang$^1$ \\
  $^1$Department of Automation, BNRist, Tsinghua University \\
  $^2$Paul G. Allen School of Computer Science and Engineering, University of Washington\\
  \texttt{\{r-lu21,zqc21\}@mails.tsinghua.edu.cn}\\
  \texttt{ssdu@cs.washington.com, gaohuang@tsinghua.edu.cn} \\
}

\begin{document}
\maketitle
\begin{abstract}
 While multitask representation learning has become a popular approach in reinforcement learning (RL) to boost the sample efficiency, the theoretical understanding of why and how it works is still limited. Most previous analytical works could only assume that the representation function is already known to the agent or from linear function class, since analyzing general function class representation encounters non-trivial technical obstacles such as generalization guarantee, formulation of confidence bound in abstract function space, etc. However, linear-case analysis heavily relies on the particularity of linear function class, while real-world practice usually adopts general non-linear representation functions like neural networks. This significantly reduces its applicability. In this work, we extend the analysis to \textit{general} function class representations. Specifically, we consider an agent playing $M$ contextual bandits (or MDPs) concurrently and extracting a shared representation function $\phi$ from a specific function class $\Phi$ using our proposed Generalized Functional Upper Confidence Bound algorithm (GFUCB). We theoretically validate the benefit of multitask representation learning within general function class for bandits and linear MDP for the first time. Lastly, we conduct experiments to demonstrate the effectiveness of our algorithm with neural net representation.
\end{abstract}

\section{Introduction}
\label{introduction}
Recently, reinforcement learning (RL) has achieved many successful applications in games \cite{berner2019dota, silver2017mastering}, robotics \cite{levine2016end}, and many other fields. However, due to the large cardinality of state space or action space in real-world problems, the large sample complexity has been a major problem for employing these RL algorithms in reality. A popular method called multitask representation learning tries to tackle this problem by extracting a shared low-dimensional representation function among multiple related tasks, then using a simple function (e.g., linear) on top of this common representation to solve each task\cite{baxter2000model,caruana1997multitask,li2010contextual}. 

Despite the empirical success for multitask representation learning, particularly in reinforcement learning because of its effectiveness in reducing sample complexity, the theoretical understanding about it is still limited. A march of works\cite{teh2017distral,taylor2009transfer,lazaric2011transfer,rusu2015policy,liu2016decoding,parisotto2015actor,hessel2019multi,arora2020provable,d'eramo2020sharing, cheng2022provable, yang2022nearly, papini2021reinforcement} give results on function approximation in bandits and RL, which permits a representation. In these frameworks, an agent is considered playing $M$ related tasks concurrently. Each task is a distinct contextual bandit or linear MDP problem \footnote{Although the name of linear MDP contains term ``linear'', it actually has infinite degrees of freedom because the representation function $\phi$ could be general non-linear function.}, and all these $M$ tasks share a common representation $\phi \in \Phi$ where $\Phi=\{\phi:\mathcal{S}\times \mathcal{A} \mapsto \mathbb{R}^k \}$ is representation function class extracting a $k$-dimensional representation vector from state-action pair. Such representation function can reduce the complexity of problem from a huge space $\mcal{S}\times\mcal{A}$ to a simple regression problem in $k$-dimensional space. The value approximation function class is defined by $\mathcal{F}=\mathcal{L}\circ \Phi$, here $\circ$ means composition and $\mathcal{L}$ means linear function, which means the value of any state-action pair $(s,a)$ is linear in its representation $\phi(s,a)$.

However, previous analyses either assume $\Phi$ is linear \cite{yang2021impact}, or assume that the agent already knows the concrete function $\phi$ \cite{hu2021near, jin2019provably}, which equivalently reduces to learning linear weight parameters. This limits their applicability, since general non-linear value estimation is ubiquitous and is the essence for the success of multitask representation learning. For instance, DQN\cite{mnih2013playing} achieves great success by employing a deep network to approximate Q-value function. Also, assuming the agent already knows a good representation function is unrealistic in practice. Therefore, we aim to extend the analysis to \textit{unknown general non-linear} representation functions. This would not only reveal the more essential benefit of multitask representation learning, but also inspire and facilitate future practice.

\subsection{Our Contribution}
The focus of previous works on linear analyses has its own reasons. The particularity of linear function could circumvent many non-trivial obstacles in analysis, which hinders previous work from from being applicable to real world scenarios. For instance, the formulation of confidence set in linear parameter space is simply an ellipsoid, and its update is straightforward via covariance matrix. More importantly, linear function class generically ensures generalization. The analysis \cite{hu2013fast, yang2019learning, lu2021power, jin2019provably} only requires the samples to span the whole input space to let the covariance matrix converge, then is able to derive uniform prediction error guarantee for the whole input space. However, generalization issue is much more complicated for general non-linear scenarios. 

In summary, our work embraces following contributions, which solves the challenges for previous works and extends the analysis for the role of representation function in more general setting. 

\textbf{Eliminate the Dependency on Linearity.} Towards general function class analysis, we adopt the idea of confidence set \cite{russo2013eluder, hu2021near}. The algorithm extends the idea of upper confidence bound and maintains a \textit{confidence set} for all the possible value estimation functions. The confidence set contains all the functions whose total empirical error at step $t$ is less than a predetermined bound $\beta_t$. As more seen data reveals more information about the environment, the confidence set will gradually shrink until converge. Therefore, our algorithm and analysis framework is applicable to general function class. 

Note that designing $\beta_t$ to achieve low regret for general function class $\Phi$ is non-trivial. We firstly determine the concrete UCB form for general function class $\beta_t(\Phi)$ and propose a straightforward algorithm called Generalized Functional Upper Confidence Bound (or GFUCB in abbreviation) for general non-linear function class approximation. We use Eluder dimension\cite{russo2013eluder} to measure the complexity of the function class $\Phi$ to give an efficient sample complexity that ensures generalization.


\textbf{Multihead Function Class.} To derive sharp regret bound for our algorithm and theoretically demonstrate the benefit of multitask representation learning, we firstly introduce multihead function class $\mcal{F}^{\otimes M}$, which is the key technical contribution of our work. The efficacy of multitask representation learning essentially originates from the shared knowledge and structure among tasks. Hence it is vital and necessary to characterize such relation between multiples tasks that the agent simultaneously learns. However, such structure is absent in previous single task work \cite{wang2020reinforcement, russo2013eluder}, and it calls for special techniques to analyze the efficiency for learning these correlated functions. 

To this end, we introduce multihead function class, namely $\mathcal{F}^{\otimes M}$ in section 4. This abstract function space captures the relation between different task functions, which concatenate the values of $(s,a)$ for all $M$ tasks together as the output. Being more compact by sharing a common backbone $\phi$, function in $\mathcal{F}^{\otimes M}$ requires much fewer samples to learn compared to $M$ independent tasks space $\mathcal{F}^{M}$. All the tasks contribute to shape a good representation, then feedback to each task for faster convergence. We formally prove that our algorithm enjoys regret bound as $\Tilde{O}\left(\sqrt{M T \operatorname{dim}_{E}(\mathcal{F})(Mk+\log \mathcal{N}(\Phi))}\right)$, where $T$ is the number of steps, $M$ is the number of tasks and $\mathcal{N}(\Phi)$ means the covering number of function space $\Phi$. We also extend the algorithm and analysis to multitask episodic RL with general value approximation under low inherent Bellman error. By simultaneously solving $M$ different but correlated MDP tasks, our method is sample-efficient with regret $\Tilde{O}\left(\sqrt{M T H \operatorname{dim}_{E}(\mathcal{F}) (Mk+\log \mathcal{N}(\Phi) + MTH\mathcal{I}^2)} \right)$ where $T$ is the number of episodes, $H$ is planning horizon and $\mathcal{I}$ denotes the inherent Bellman error. 

To the best of our knowledge, this is the first provably sample efficient algorithm for general representation function bandits and linear MDP. It is comparable to the most optimal regret bound when $\Phi$ is specialized to linear representation, and is better than the bounds which solve each task independently. This also theoretically explains how multitask representation learning reduces sample complexity. Essentially, the joint training for the shared representation function helps accelerate the convergence of the common backbone by having more samples from all the tasks.



\textbf{Empirical Value.} Finally, we conduct experiments to verify our theoretical result. We design a neural network based bandit environment and implement the GFUCB algorithm. Experimental results corroborate the effect of multitask representation learning in boosting sample efficiency in non-linear bandits. For the first time, the efficacy of the general representation algorithm proposed in theoretical analysis is validated in a proof-of-concept experiment. 

\section{Related Work}
In the supervised learning setting, a line of works have been done on multitask learning and representation learning with various assumptions \cite{baxter2000model,du2017hypothesis,ando2005framework,ben2003exploiting,maurer2006bounds,cavallanti2010linear,maurer2016benefit,du2020few,tripuraneni2020provable}.
These results assumed that all tasks share a joint representation function.
It is also worth mentioning that \cite{tripuraneni2020provable} gave the method-of-moments estimator and built the confidence ball for the feature extractor, which inspired our algorithm for the infinite-action setting.

The benefit of representation learning has been studied in sequential decision-making problems, especially in RL domains. Arora et al. \cite{arora2020provable} proved that representation learning could reduce the sample complexity of imitation learning.
D'eramo et al. \cite{d'eramo2020sharing} showed that representation learning could improve the convergence rate of the value iteration algorithm. Both require a probabilistic assumption similar to that in \cite{maurer2016benefit}, and the statistical rates are of similar forms as those in \cite{maurer2016benefit}. Following these works, we study a special class of MDP called Linear MDP. Linear MDP \cite{yang2019sample,jin2019provably} is a popular model in RL, which uses linear function approximation to generalize large state-action space. \cite{zanette2020learning} extends the definition to low inherent Bellman error (or IBE in short) MDPs. This model assumes that both the transition and the reward are near-linear in given features. 

Recently, Yang et al. \cite{yang2021impact} showed multitask representation learning reduces the regret in linear bandits, using the framework developed by Du et al. \cite{du2020few}. Moreover, some works \cite{hu2021near, lu2021power, jin2019provably} proved results on the benefit of multitask representation learning RL with generative model or linear representation function. However, these works either restrict the representation function class to be linear, or the representation function is known to agent. This is unrealistic in real world practice, which limits these works' meaning.

The most relevant works that need to be mentioned is general function class value approximation for bandits and MDPs. Russo et al. \cite{russo2013eluder} first proposed the concept of eluder dimension to measure the complexity of a function class and gave a regret bound for general function bandits using this dimension. Wang et al. \cite{wang2020reinforcement} further proved that it can also be adopted in MDP problems. Dong et al. \cite{dong2021provable} extended the analysis with sequential Rademacher complexity. Inspired by these works, we adopt eluder dimension and develop our own analysis. But it should be pointed out that all those works focus on single task setting, which give a provable bound for just one single MDP or bandit problem. They lack the insight for why simultaneously dealing with multiple distinct but correlated tasks is more sample efficient. Our work aim to establish a framework to explain this. By considering locating the ground truth value function in multihead function space $\mathcal{F}^{\otimes M}$ (see detailed definition in \hyperref[resbandit]{section 4}), we are able to theoretically explain the main reason for the boost of sample efficiency. Informally speaking, the shared feature extraction backbone $\phi$ receives samples from all the tasks, therefore accelerating the convergence for every single task compare with solving them separately.

\section{Preliminaries}
\label{prelim}
\subsection{Notations}
We use $[n]$ to denote the set $\{1,2,\hdots,n\}$ and $\langle \cdot,\cdot\rangle$ to denote the inner product between two vectors. We use $f(x) = O(g(x))$ to represent $f(x) \leq C\cdot g(x)$ holds for any $x>x_0$ with some $C>0$ and $x_0>0$. Ignoring the logarithm term, we use $f(x) = \tilde{O}(g(x))$ .
\subsection{Multitask Contextual Bandits}
We first study multitask representation learning in contextual bandits. Each task $i\in[M]$ is associated with an unknown function $f^{(i)}\in \mathcal{F}$ from certain function class $\mathcal{F}$. At each step $t\in[T]$, the agent is given a context vector $C_{t,i}$ from certain context space $\mathcal{C}$ and a set of actions $\mathcal{A}_{t,i}$ selected from certain action space $\mathcal{A}$ for each task $i$. The agent needs to choose one action $A_{t,i}\in \mathcal{A}_{t,i}$, and then receives a reward as $R_{t,i} = f^{(i)}(C_{t,i}, A_{t,i}) + \eta_{t,i}$, where $\eta_{t,i}$ is the random noise sampled from some i.i.d. distribution. The agent's goal is to understand function $f^{(i)}$ and maximize the cumulative reward, or equivalently, minimize the total regret from all $M$ tasks in $T$ steps defined as below.
$$
\operatorname{Reg}(T) \stackrel{\text { def }}{=} \sum_{t=1}^{T} \sum_{i=1}^{M}\left( f^{(i)}(C_{t,i},A_{t,i}^{\star}) - f^{(i)}(C_{t,i},A_{t,i}) \right),
$$
where $A_{t,i}^{\star} = \arg\max_{A\in\mcal{A}_{t,i}} f^{(i)}(C_{t,i}, A)$ is the optimal action with respect to context $C_{t,i}$ in task $i$. 

\subsection{Multitask MDP}
Going beyond contextual bandits, we also study how this shared low-dimensional representation could benefit the sequential decision making problem like Markov Decision Process (MDP). In this work, we study undiscounted episodic finite horizon MDP problem. Consider an MDP $\mcal{M}=(\mcal{S}, \mcal{A}, \mcal{P}, r, H)$, where $\mcal{S}$ is the state space, $\mcal{A}$ is the action space, $\mcal{P}$ is the transition dynamics, $r(\cdot, \cdot)$ is the reward function and $H$ is the planning horizon. The agent starts from an initial state $s_1$ which can be either fixed or sampled from a certain distribution, then interacts with environment for $H$ rounds. In the single task framework, at each round (also called level) $h$, the agent needs to perform an action $a_h$ according to a policy function $a_h=\pi_h(s_h)$ . Then the agent will receive a reward $R_h(s_h, a_h) = r(s_h, a_h) + \eta_{h}$ where $\eta_h$ again is the noise term. The environment then transits the state from $s_{h}$ to $s_{h+1}$ according to distribution $\mcal{P}(\cdot | s_h, a_h)$. The estimation for action value function given following action policy $\pi$ is defined as $Q_h^{\pi}(s_h, a_h) = r(s_h,a_h)+\mathbb{E}\left[ \sum_{t=h+1}^H R_t(s_t, \pi_t(s_t)) \right]$, and state value function is defined as $V_h^{\pi}(s_h) = Q_h^{\pi}(s_h, \pi_h(s_h))$. Note that there always exists a deterministic optimal policy $\pi^{\star}$ for which $V_h^{\pi^{\star}}(s)=\max_{\pi} V_h^{\pi}(s)$ and $Q_h^{\pi^{\star}}(s,a) = \max_{\pi} Q_h^{\pi}(s,a)$, we will denote them as $V_h^{\star}(s)$ and $Q_h^{\star}(s,a)$ for simplicity. 

In the multitask setting, the agent gets a batch of states $\{s_{h,t}^{(i)}\}_{i=1}^M$ simultaneously from $M$ different MDP tasks $\{\mcal{M}^{(i)}\}_{i=1}^M$ at each round $h$ in episode $t$, then performs a batch of actions $\{\pi_t^i(s_{h,t}^{(i)})\}_{i=1}^M$ for each task $i\in[M]$. Every $H$ rounds form an episode, and the agent will interact with the environment for totally $T$ episodes. The goal for the agent is minimizing the regret defined as 
\begin{align*}
    \operatorname{Reg}(T) = \sum_{t=1}^T \sum_{i=1}^M V_1^{(i)\star} \left( s_{1,t}^{(i)} \right) - V_1^{\pi_t^i}\left( s_{1,t}^{(i)} \right),
\end{align*}
where $V_1^{(i)\star}$ is the optimal value of task $i$ and $s_{1,t}^{(i)}$ is the initial state for task $i$ at episode $t$.

To let representation function play a role, it is assumed that all tasks share the same state space $\mcal{S}$ and action space $\mcal{A}$. Moreover, there exists a representation function $\phi:\mcal{S}\times\mcal{A} \mapsto \mathbb{R}^k$ such that action and state value function of all tasks $\mcal{M}^{(i)}$ is always (approximately) linear in this representation. For example, given a representation function $\phi$, the action value approximation function at level $h$ is parametrized by a vector $\bsyb{\theta}_h \in \mathbb{R}^k$ as $Q_h[\phi, \bsyb{\theta}_h] \stackrel{\text { def }}{=} \langle \phi(s,a), \bsyb{\theta}_h \rangle$, similar for $V_h [\phi, \bsyb{\theta}_h](s) \stackrel{\text { def }}{=} \max_{a} \langle \phi(s,a), \bsyb{\theta}_h \rangle$. We denote all such action value functions as $\mcal{Q}_h=\{Q_h[\phi, \bsyb{\theta}_h]:\phi\in\Phi,\bsyb{\theta}_h\in\mathbb{R}^k\}$, also value function approximation space as $\mcal{V}_h=\{V_h[\phi, \bsyb{\theta}_h]:\phi\in\Phi,\bsyb{\theta}_h\in\mathbb{R}^k\}$. Each task $\mcal{M}^{(i)}$ is a linear MDP, which means $\mcal{Q}_h$ is always approximately close under Bellman operator $\mcal{T}_h(Q_{h+1})(s,a) \stackrel{\text { def }}{=} r_h(s,a)+\mathbb{E}_{s'\sim \mcal{P}_h(\cdot|s,a)} \max_{a'}Q_{h+1}(s',a')$.

\textbf{Linear MDP Definition.} \textit{A finite horizon MDP $\mcal{M}=(\mcal{S}, \mcal{A}, \mcal{P}, r, H)$ is a linear MDP, if there exists a representation function $\phi:\mcal{S}\times\mcal{A} \mapsto \mathbb{R}^k$ and its induced value approximation function class $\mcal{Q}_h, h\in[H]$, such that the inherent Bellman error}\cite{zanette2020learning}
$$
\mathcal{I}_{h} \stackrel{\text { def }}{=} \sup _{Q_{h+1} \in \mathcal{Q}_{h+1}} \inf _{Q_{h} \in \mathcal{Q}_{h}} \sup _{s \in \mathcal{S}, a \in \mathcal{A}}\left|\left(Q_{h}-\mathcal{T}_{h}\left(Q_{h+1}\right)\right)(s, a)\right|,
$$
\textit{is always smaller than some small constant $\mcal{I}$.}

The definition essentially assumes that for any Q-value approximation function $Q_{h+1} \in \mcal{Q}_{h+1}$ at level $h+1$, the Q-value function $Q_h$ at level $h$ induced by it can always be closely approximated in class $\mcal{Q}_{h}$, which assures the accuracy through sequential levels.

\subsection{Eluder Dimension}
To measure the complexity of a general function class $f$, we adopt the concept of eluder dimension \cite{russo2013eluder}. First, define $\epsilon$-dependence and independence.

\textbf{Definition 1 ($\epsilon$-dependent).} \textit{An input $x$ is \textit{$\epsilon$-dependent} on set $X=\{x_1, x_2, \hdots, x_n\}$ with respect to function class $\mcal{F}$, if any pair of functions $f, \tilde{f} \in \mcal{F}$ satisfying $\sqrt{ \sum_{i=1}^n (f(x_i) - \tilde{f}(x_i))^2} \leq \epsilon$ also satisfies $| f(x) - \tilde{f}(x) | \leq \epsilon$. Otherwise, we call action $x$ to be $\epsilon$-independent of data set $X$.}

Intuitively, $\epsilon$-dependence captures the exhaustion of interpolation flexibility for function class $\mcal{F}$. Given an unknown function $f$'s value on set $X=\{x_1, x_2, \hdots, x_n\}$, we are able to pin down its value on some particular input $x$ with only $\epsilon$-scale prediction error. 

\textbf{Definition 2 ($\epsilon$-eluder dimension).} \textit{The \textit{$\epsilon$-eluder} dimension $\operatorname{dim}_E(\mcal{F}, \epsilon)$ is the maximum length for a sequence of inputs $x_1,x_2,\hdots x_d\in \mcal{X}$, such that for some $\epsilon'\geq \epsilon$, every element is $\epsilon'$-independent of its predecessors.}

This definition is similar to the definition of the dimensionality of a linear space, which is the maximum length of a sequence of vectors such that each one is linear independent to its predecessors. For instance, if $\mcal{F}=\{f(x):\mathbb{R}^d \mapsto \mathbb{R},f(x)=\theta^{\top} x\}$, we have $\operatorname{dim}_E(\mcal{F},\epsilon) = O(d \log 1/\epsilon)$ since any $d$ linear independent input's estimated value can fully describe a linear mapping function. We also omit the $\epsilon$ and use $\operatorname{dim}_E(\mcal{F})$ when it only has a logarithm dependent term on $\epsilon$. 

\section{Main Results for Contextual Bandits}
In this section, we will present our theoretical analysis on the proposed GFUCB algorithm for contextual bandits.
\label{resbandit}
\subsection{Assumptions}
This section will list the assumptions that we make for our analysis. The main assumption is the existence of a shared feature extraction function from class $\Phi=\{\phi:\mcal{C}\times \mcal{A}\mapsto \mathbb{R}^k\}$ that any task's value function is linear in this $\phi$.

\textbf{Assumption 1.1 (Shared Space and Representation)}
\textit{All the tasks share the same context space $\mcal{C}$ and action space $\mcal{A}$. Also, there exists a shared representation function $\phi\in\Phi$ and a set of $k$-dimensional parameters $\{\bsyb{\theta}_i\}_{i=1}^M$ such that each $f^{(i)}$ has the form $f^{(i)}(\cdot,\cdot) = \langle\phi(\cdot,\cdot),\bsyb{\theta}_i\rangle$.}

Following standard regularization assumptions for bandits \cite{hu2021near, yang2021impact}, we make assumptions on noise distribution and function parameters.

\textbf{Assumption 1.2 (Conditional Sub-Gaussian Noise)}
\textit{Denote $ \mcal{H}_{t,i} = \sigma(C_{1,i}, A_{1,i}, \hdots, C_{t,i}, A_{t,i}) $ to be the $\sigma$-field summarizing the history information available before reward $R_{t,i}$ is observed for every task $i\in[M]$. We have $\eta_{t,i}$ is sampled from a 1-Sub-Gaussian distribution, namely $\mathbb{E}\left[ \exp(\lambda \eta_{t,i}) \mid \mcal{H}_{t,i} \right] \leq \exp \left( \frac{\lambda^2}{2}\right) $ for $\forall \lambda \in \mathbb{R}$}

\textbf{Assumption 1.3 (Bounded-Norm Feature and Parameter)} 
\textit{We assume that the parameter $\bsyb{\theta}_i$ and the feature vector for any context-action pair $(C, A)\in \mcal{C}\times \mcal{A}$ is constant bounded for each task $i \in [M]$, namely $\| \bsyb{\theta}_i \|_2 \leq \sqrt{k}$ for $\forall i\in[M]$ and $\| \phi(C, A) \|_2 \leq 1$ for $\forall C\in \mcal{C}, A \in \mcal{A}$.}
    
Apart from these assumptions, we add assumption to measure and constrain the complexity of value approximation function class $\mcal{F} = \mcal{L} \circ \Phi$. 
    
\textbf{Assumption 1.4 (Bounded Eluder Dimension).} \textit{We assume that function class $\mcal{F}$ has bounded Eluder dimension $d$, which means for any $\epsilon$, $\operatorname{dim}_{E}(\mcal{F}, \epsilon) = \tilde{O}(d)$.}
\subsection{Algorithm Details}
    \begin{algorithm}[ht]
    \label{alg:alg1}
        \caption{Generalized Functional UCB Algorithm}
        \begin{algorithmic}[1]
        \FOR{step $t:1 \to T$}
            \STATE Compute $\mcal{F}_t$ according to (\hyperref[equ:optstar]{$*$}) 
            \STATE Receive contexts $C_{t,i}$ and action sets $\mcal{A}_{t,i}$, $i \in [M]$
            \STATE $
                f_t, A_{t,i} = \mathop{\mathrm{arg max}}_{f\in \mcal{F}_t,\ A_{i} \in \mcal{A}_{t,i}} \sum_{i=1}^M f^{(i)}(C_{t,i}, A_i)
            $
            \STATE Play $A_{t,i}$ for task i, and get reward $R_{t,i}$ for $i\in[M]$.
        \ENDFOR
        \end{algorithmic}
    \end{algorithm}
The details of the algorithm is in \hyperref[alg:alg1]{Algorithm 1}. At each step $t$, the algorithm first solves the optimization problem below to get the empirically optimal solution $\hat{f}_{t}$ that best predicts the rewards for context-input pairs seen so far.
\begin{align*}
    \hat{f}_{t} \gets \mathop{\mathrm{argmin}}_{f\in\mcal{F}^{\otimes M}} \sum_{i=1}^M \sum_{k=1}^{t-1} \left( f^{(i)}( C_{k,i}, A_{k,i}) - R_{k,i} \right)^2
\end{align*}
Here we abuse the notation of $\mcal{F}^{\otimes M}$ as $\mcal{F}^{\otimes M}=\left\{f=\left( f^{(1)}, \hdots, f^{(M)} \right): f^{(i)}(\cdot) = \phi(\cdot)^{\top} \bsyb{w}_i \in \mcal{F} \right\}$ to denote the M-head prediction version of $\mcal{F}$, parametrized by a shared representation function $\phi(\cdot)$ and a weight matrix $\bsyb{W}=[\bsyb{w}_1,\hdots,\bsyb{w}_M] \in \mathbb{R}^{k\times M} $. We use $f^{(i)}$ to denote the $i_{th}$ head of function $f$ which specially serves for task $i$.


After obtaining $\hat{f}_{t}$, we maintain a functional confidence set $\mcal{F}_t \subseteq \mcal{F}^{\otimes M}$ for possible value approximation functions 
\begin{align*}
\label{equ:optstar}
\mathcal{F}_{t} \stackrel{\text { def }}{=} \Bigg\{ &f \in \mcal{F}^{\otimes M} :  \left\| \hat{f}_{t} - f \right\|^{2}_{2,E_t} \leq \beta_t, | f^{(i)}(\bsyb{x}) | \leq 1, \forall \bsyb{x}\in \mcal{C}\times\mcal{A},i\in[M] \Bigg\} \tag{$*$}
\end{align*}

Here, for the sake of simplicity, we use 
$\left\| \hat{f}_{t} - f \right\|^{2}_{2,E_t} = \sum_{i=1}^M \sum_{k=1}^{t-1} \left( \hat{f}_t^{(i)} (\bsyb{x}_{k,i}) - f^{(i)}(\bsyb{x}_{k,i}) \right)^2$
to denote the empirical 2-norm of function $\hat{f}_t - f = \left( \hat{f}_t^{(1)}-f^{(1)}, \hdots, \hat{f}_t^{(M)}-f^{(M)} \right)$. Basically, ($*$) contains all the functions in $\mcal{F}^{\otimes M}$ whose value estimation difference on all collected context-action pairs $\bsyb{x}_{k,i}=(C_{k,i}, A_{k,i})$ compared with empirical loss minimizer $\hat{f}_t$ does not exceed a preset parameter $\beta_t$. We show that with high probability, the real value function $f_{\theta}$ is always contained in $\mcal{F}_t$ when $\beta_t$ is carefully chosen as $\tilde{O}(Mk+\log \left(\mcal{N}(\Phi, \alpha, \|\cdot\|_{\infty})\right)$, where $\mcal{N}(\mcal{F}, \alpha, \|\cdot\|_{\infty})$ is the $\alpha$-covering number of function class $\Phi$ in the sup-norm $\|\phi \|_{\infty} = \max_{\bsyb{x} \in\mcal{S}\times \mcal{A}} \| \phi(\bsyb{x})\|_2$ and $\alpha$ is set to be a small number as $\frac{1}{kMT}$ (see detailed definition and proof in Lemma 1).

For the action choice, our algorithm follows \textit{OFUL}, which estimates each action value with the most optimistic function value in our confidence set $\mcal{F}_{t}$, and chooses the action whose optimistic value estimation is the highest. In the multitask setting, we choose one action from each task to form an action tuple $(A_1,A_2,\hdots, A_M)$ such that the summation of the optimistic value estimation $\sum_{i=1}^M f^{(i)}(C_{t,i}, A_i)$ is maximized by some function $f\in\mcal{F}_t$.

\textbf{Intractability.} Some may have concerns on the intractability of building the confidence set ($*$) and solving the optimization problem to get $\hat{f}_t, f_t, A_{t,i}$. The solution comes as two folds. From the theoretical perspective, since the focus of problem is sample complexity rather than computational complexity, a computational oracle can simply be assumed to give the solution of the optimization. This is the common practice for theoretical works \cite{jin2021bellman, sun2018model, agarwal2014taming, jiang2017contextual} in order to focus on the sample complexity analysis. From empirical perspective, there are great chances to optimize it with gradient methods. For example, solving $\hat{f}_t$ is a standard empirical risk minimization problem, and can be effectively solved with gradient methods \cite{du2019gradient}. As for $f_{t}$ and $A_{t,i}$, note that it is not necessary to explicitly build the confidence set $\mcal{F}_t$ by listing all the candidates. The approximation algorithm just need to search within the confidence set via gradient method to optimize objective $\sum_{i=1}^M f^{(i)}(C_{t,i}, A_i)$. The start point is $\hat{f}_t$, and the algorithm knows that it approaches the border of $\mathcal{F}_t$ when $\|\hat{f}_t-f\|_{2,E_t}^2$ approaches $\beta_t$. The details of implementation are in section 6.

\textbf{Mechanism.} GFUCB algorithm solves the exploration problem in an implicit way. For a context-action pair $\bsyb{x}=(C,A)$ in task $i$ which has not been fully understood and explored yet, the possible value estimation $f^{(i)}(\bsyb{x})$ will vary in large range with regard to constraint $\|f-\hat{f}_t\|_{2,E_t}^2 \leq \beta_t$. This is because within $\mcal{F}_t$ there are many possible function value on this $\bsyb{x}$ while agreeing on all past context-action pairs' value. Therefore, the optimistic value $f^{(i)}(\bsyb{x})$ will become high by getting a significant implicit bonus, encouraging the agent to try such action $A$ under context $C$, which achieves natural exploration. 

The reduction of sample complexity is achieved through joint training for function $\phi$. If we solve these tasks independently, the confidence set width $\beta_t$ is at scale $M\log \left(\mcal{N}(\Phi, \alpha, \|\cdot\|_{\infty})\right)$ because it needs to cover $M$ representation function space respectively. By involving $\phi$ in the prediction for all tasks, our algorithm reduces the size of confidence set by $M$ times, since now the samples from all the tasks can contribute to learn the representation $\phi$. Usually $\log \left(\mcal{N}(\Phi, \alpha, \|\cdot\|_{\infty}) \right)$ is much greater than $k$ and $M$, hence our confidence set shrinks at a much faster speed. This explains how GFUCB achieves lower regret, since the sub-optimality at each step $t$ is proportional to the confidence set width $\beta_t$ when real value function $f_{\theta}\in\mcal{F}_t$.

\subsection{Regret Bound}
Based on the assumptions above, we have the regret guarantee as below.

\textbf{Theorem 1.} \textit{Based on assumption 1.1 to 1.4, denote the cumulative regret in $T$ steps as $\operatorname{Reg}(T)$, with probability at least $1-\delta$ we have }$\operatorname{Reg}(T) = \tilde{O} \left( \sqrt{M d T (Mk + \log\mcal{N}(\Phi, \alpha_{T}, \| \cdot \|_{\infty})) } \right).$

Here, $d:=\operatorname{dim}_{E}(\mcal{F}, \alpha_{T})$ is the Eluder dimension for value approximation function class $\mcal{F} = \mcal{L} \circ \Phi$, and $\alpha_T$ is discretization scale which only appears in logarithm term thus omitted. The detailed proof is left in appendix. 

To the best of knowledge, this is the first regret bound for general function class representation learning in contextual bandits. To get a sense of its sharpness, note that when $\Phi$ is specialized as linear function class as $\Phi=\{\phi(x)=\bsyb{Bx}, \bsyb{B}\in\mathbb{R}^{k\times d}\}$, we have $\log\mcal{N}(\Phi, \alpha_{T}, \| \cdot \|_{\infty}) = \tilde{O}(dk)$ and $\operatorname{dim}_{E}(\mcal{F})=d$, then our bound is reduced to $\tilde{O}(M\sqrt{dTk}+d\sqrt{MTk})$, which is the same optimal as the current best provable regret bound for linear representation class bandits in \cite{hu2021near}.

\section{Main Results for MDP}
\label{res_MDP}
\subsection{Assumptions}
For multitask Linear MDP setting, we adopt Assumption 3 from \cite{hu2021near} which generalizes the inherent Bellman error \cite{zanette2020learning} to multitask setting. 

\textbf{Assumption 2.1 (Low IBE for multitask)} \textit{Define multi-task IBE is defined as }
$$
\mathcal{I}_{h}^{\text {mul }} \stackrel{\text { def }}{=} \sup_{\left\{Q_{h+1}^{(i)}\right\}_{i=1}^{M} \in \mathcal{Q}_{h+1}} \inf_{\left\{Q_{h}^{(i)}\right\}_{i=1}^{M} \in \mathcal{Q}_{h}} \sup_{s \in \mathcal{S}, a \in \mathcal{A}, i \in[M]} \left|\left(Q_{h}^{(i)}-\mathcal{T}_{h}^{(i)}\left(Q_{h+1}^{(i)}\right)\right)(s, a)\right|.
$$
\textit{We have $\mcal{I} \stackrel{\mathrm{def}}{=} \sup_{h} \mathcal{I}_{h}^{\mathrm{mul }}$ is small  for all $\mcal{Q}_h$, $h\in[H]$.}

Assumption 2.1 generalize low IBE to multitask setting. It assumes that for every task $i\in [M]$, its Q-value function space is always close under Bellman operator.

\textbf{Assumption 2.2 (Parameter Regularization)} \textit{We assume that}
\begin{itemize}
\setlength{\itemsep}{0pt}
\setlength{\parsep}{0pt}
\setlength{\parskip}{3pt}
    \item \textit{$\|\phi(s,a)\|\leq 1$, $0 \leq Q_h^{\pi}(s,a) \leq 1$ for $\forall (s,a)\in \mcal{S}\times \mcal{A}, h\in[H], \forall \pi$.}
    \item \textit{There exists a constant $D$ such that for any $h\in[H]$ and $\bsyb{\theta}_h^{(i)}$, it holds that $\|\bsyb{\theta}_h^{(i)}\|_2 \leq D$.}
    \item \textit{For any fixed $\left\{Q_{h+1}^{(i)} \right\}_{i=1}^M \in \mcal{Q}_{h+1}$, the random noise $z_{h}^{(i)} \stackrel{\mathrm{def}}{=} R_{h}^{(i)}(s,a) + \max_{a} Q_{h+1}^{(i)}(s',a) - \mcal{T}_h^{(i)} \left(Q_{h+1}^{(i)}\right)(s,a)$ is bounded in $[-1,1]$ and is always independent to all other random variables for $\forall (s,a)\in\mcal{S}\times\mcal{A}, h\in[H], i\in[M]$.}
\end{itemize}
These assumptions are widely adopted in linear MDP analytical works \cite{zanette2020learning, hu2021near, lu2021power}, which regularizes the parameter, feature, and noise scale. Again we add bounded Eluder dimension constraint for the Q-value estimation class. 

\textbf{Assumption 2.3 (Bounded Eluder Dimension).} \textit{We assume that function class $\mcal{Q}_h$ has bounded Eluder dimension $d$ for any $h\in[H]$.}
\subsection{Algorithm Details}
    \begin{algorithm}[ht]
    \label{alg:alg2}
        \caption{multitask Linear MDP Algorithm}
        \begin{algorithmic}[1]
        \FOR{episode $t:1 \to T$}
            \STATE $Q_{H+1}^{(i)}=0, i\in[M]$ 
            \FOR{$h: H \to 1$}
                \STATE $\hat{\phi}_{h,t},\hat{\bsyb{\theta}}_{h,t}^{(i)} \gets $ solving (1)
                \STATE $Q_{h}^{(i)}(\cdot, \cdot) = \hat{\phi}_{h,t}(\cdot,\cdot)^{\top} \hat{\bsyb{\theta}}_{h,t}^{(i)}, V_h^{(i)}(\cdot) = \max_{a} Q_{h}^{(i)}(\cdot, a)$
            \ENDFOR
            \FOR {$h: 1 \to H$}
                \STATE Compute $\mcal{F}_{h,t}$ according to \hyperref[prf:lemma4]{Lemma 4}
                \STATE Receive states $\left\{ s_{h,t}^{(i)} \right\}_{i=1}^M$, 
                        $\tilde{f}_{h,t}, a_{h,t}^{(i)} = \mathop{\mathrm{arg max}}_{f\in \mcal{F}_{h,t},a^{(i)}\in\mcal{A}} \sum_{i=1}^M f^{(i)} \left( s_{h,t}^{(i)} , a^{(i)} \right)$
                    \STATE Play $a_{h,t}^{(i)}$ and get reward $R_{h,t}^{(i)}$ for task $i\in[M]$.
            \ENDFOR
        \ENDFOR
        \end{algorithmic}
    \end{algorithm}

The algorithm for multitask linear MDP is similar to contextual bandits as above. The optimization problem in line 4 of \hyperref[alg:alg2]{Algorithm 2} is finding the empirically best solution for Q-value estimation at level $h$ in episode $t$ as below
\begin{align*}
    \hat{\phi}_{h,t}, \hat{\bsyb{\Theta}}_{h,t} \gets&\mathop{\mathrm{argmin}}_{\phi\in\Phi, \bsyb{\Theta}=[\bsyb{\theta}^{(1)},\hdots,\bsyb{\theta}^{(M)}]} \mcal{L}(\phi, \bsyb{\Theta}) \tag{1} \\
    s.t.\quad& \|\bsyb{\theta}^{(i)}\| \leq D, \forall i \in[M]\\
    & 0\leq \phi(s,a)^{\top} \bsyb{\theta}_i \leq 1, \forall (s,a)\in \mcal{S}\times\mcal{A}, i\in[M],
\end{align*}
where $\mcal{L}(\phi, \bsyb{\Theta})$ is the empirical loss function defined as
\begin{align*}
\sum_{i=1}^M \sum_{j=1}^{t-1} \left( \phi\left(s^{(i)}_{h,j}, a^{(i)}_{h,j} \right)^{\top} \bsyb{\theta}^{(i)} - R_{h,j}^{(i)} - V_{h+1}^{(i)} \left( s^{(i)}_{h+1,j} \right) \right)^2.
\end{align*}
The framework of our work resembles LSVI \cite{jin2019provably} and \cite{lu2021power} which learns the Q-value estimation in a reverse order, at each level $h$, the algorithm uses just-learned value estimation function $V_{h+1}$ to build the regression target value as $R_{h,j}^{(i)} + V_{h+1}^{(i)} \left( s^{(i)}_{h+1,j} \right)$ and find empirically best estimation $\hat{f}_{h,t}^{(i)} =\hat{\phi}_{h,t}^{\top} \hat{\bsyb{\theta}}_{h,t}^{(i)}$ for each task $i\in[M]$. The optimistic value estimation of each action is again searched within confidence set $\mcal{F}_{h,t}$ which centered at $\hat{f}_{h,t}$ and shrinks as the constraint $\|f-\hat{f}_{h,t}\|^2_{2,E_t} \leq \beta_{t}$ becomes increasingly tighter. Note that the contextual bandit problem can be regarded as a 1-horizon MDP problem without transition dynamics, and our framework at each level $h$ is indeed a copy of procedures in Algorithm 1.
\subsection{Regret Bound}
Based on assumptions 2.1 to 2.3, we prove that our algorithm enjoys a regret bound guaranteed by the following theorem. Detailed proof is left in appendix.

\textbf{Theorem 2.} \textit{Based on assumption 2.1 to 2.3, denote the cumulative regret in $T$ episodes as $\operatorname{Reg}(T)$, we have the following regret bound for $\operatorname{Reg}(T)$ holds with probability at least $1-\delta$ for Algorithm 2}
\begin{equation*}
\setlength{\abovedisplayskip}{2ex}
    \tilde{O}\left( MH\sqrt{Tdk} + H\sqrt{M T d \log\mcal{N}(\Phi, \alpha)} + MHT\mcal{I}\sqrt{d} \right),
\setlength{\belowdisplayskip}{2ex}
\end{equation*}
\textit{where $\alpha$ is discretization scale smaller than $\frac{1}{kMT}$.}

\textbf{Remark.} Compared with naively executing single task general value function approximation algorithm \cite{wang2020reinforcement} for $M$ tasks, whose regret bound is $\tilde{O}(MHd\sqrt{T\log\mcal{N}(\Phi)})$, to achieve same average regret, our algorithm outperforms this naive algorithm with a boost of sample efficiency by $\tilde{O}(Md)$. This benefit mainly attributes to learning in function space $\mcal{F}^{\otimes M}=\mcal{L}^M\circ \Phi$ instead of $\mcal{F}^M = (\mcal{L}\circ \Phi)^M$, the former is more compact and requires much less samples to learn.

\section{Experiments}
\label{exp}
To validate our theoretical findings, we conduct experiments on a non-linear neural network bandits. Note that it is a proof-of-concept experiment. Our main purpose is to realize the GFUCB algorithm and check its efficacy but \textit{not} to beat sophisticated real-world algorithms. The point to demonstrate is that sample efficiency of GFUCB is scalable to the number of tasks and better than naive exploration.
\subsection{Task Design}
To test the efficacy of our algorithm, we use the MNIST dataset \cite{deng2012mnist} to build a bandit problem that involves non-linear value approximation. The reward function of the bandit environment maps the same digit into the same base reward $r_b$, which ranges from 0 to 1, plus a noise $\eta_h$ sampled from a zero-mean Gaussian with a standard deviation of 0.01. At every round, each task will present the agent a context $C$ consists of $K$ different digit images and ask the agent to take action as an integer $j\in[K]$ meaning which image to choose, then return the reward according to the agent's choice.

For the multitask setting, we construct $M$ different tasks using different digit-to-reward mappings $\sigma_{i}: \{0,\hdots,9\} \mapsto [0,1], i\in[M]$, where $\sigma_i(k)$ will give a unique reward for all images of digit $k$ in task $i$. Different tasks have different reward mapping function $\sigma_i(\cdot)$. By designing the environment this way, it requires to learn a common representation $\phi$ to recognize digits for different tasks. 


\subsection{Implementation Details}
We use a simple CNN as our feature extraction function $\phi$, which takes a digit image as input and outputs a 10-dimensional normalized vector as representation. It consists of two 3x3 convolution layers and two fully-connected layers, followed by ReLU activation and a normalization procedure.

The biggest challenge for implementation is how to solve a complex optimization problem in general functional space. In principle, finding parameters for a neural network to achieve the (near) minimal empirical error is an NP-Hard problem. To solve this issue, we use a gradient-based method to approximately find a local-optimal solution. For finding the empirically best $\hat{f}_{t}$, we use Adam with $lr=1e-3$ to train for sufficiently long steps; in our setting, it is set to be 200 epochs at every step $t$, to ensure that the training loss is sufficiently low. 

The next major challenge is estimating the optimistic value for each action within the abstract function set $\mcal{F}_t$. To tackle this problem, we enumerate all possible action tuples $\{A_{i}\}_{i=1}^M$ and then solve the equivalent optimization below to compute its optimistic estimated value
$$
     \max_{f\in \mcal{F}_{t}} \sum_{i=1}^M f^{(i)}(C_{t,i}, A_{i})\quad s.t.\quad  \left\| f - \hat{f}_t \right\|_{2,E_t}^2 \leq \beta_t.
$$
Still, this is a complicated optimization problem within an abstract function set. Inspired by the Lagrangian operator, we transform it into an unconstrained optimization problem minimizing loss function $
    \ell(f) = - \sum_{i=1}^M f^{(i)}(A_i) + \lambda \cdot \max (0, \| \hat{f}_{t} - f \|^{2}_{2,E_t}- B_{t} )
$, where $\lambda$ is a hyperparameter to be determined, in our algorithm we set it to be $\lambda=30$ by empirical search. Also $B_t=a\log(b\cdot t+c)$ is an approximation for $\beta_t$ since $\beta_t$ includes $\mcal{N}(\Phi, \alpha)$ which is intractable to be exactly computed, we found $(a,b,c)=(0.4,0.5,2)$ to be a good parameter of UCB in single task. We use SGD with a small learning rate ($5e-4$) to finetune the model $\hat{f}_t$ for 200 iterations to optimize $\ell(f)$. 

The basic intuition is that, through optimizing $\ell(f)$, the algorithm will try to maximize function value $\sum_{i=1}^M f^{(i)}(A_i)$. And as long as $f$ satisfies $\| \hat{f}_{t} - f \|^{2}_{2,E_t}\leq B_{t}$, such constraint will not appear in the loss term, thus has no effect on optimization. When $f$ comes to the border of $\mcal{F}_t$, where $\| \hat{f}_{t} - f\|^{2}_{2,E_t}$ approaches $B_t$, the second term adds regularization term to the loss as punishment, preserving $\| \hat{f}_{t} - f \|^{2}_{2,E_t}$ at a near-constant level around $B_t$. So we can approximately simulate the optimistic value estimating procedure via searching in the neighborhood of $\hat{f}_t$.

\subsection{Connection to Algorithm 1}
The main difference between our practical version algorithm and the theoretical one is that we did not list out all the functions in the whole confidence set $\mathcal{F}_t$ explicitly, but just use gradient-based method to implicitly search within a very small fraction of $\mathcal{F}_t$ with heuristics. Getting a candidate within the confidence set is much easier and tractable than rigorously exhausting all functions in $\mathcal{F}_t$ to optimize. We can start from the parameter of $\hat{f}_t$ and use gradient method to approximately find $f_t$ and $A_{t,i}$.

Another difference is we do not rigorous compute $\beta_t$ which involves $\mathcal{N}(\Phi)$, but directly determine a parametrized function form. Rigorously speaking, our tuned value of $\beta_t$ is much smaller than the theoretical guaranteed ones, so all the candidate functions that we search along the trajectory of gradient method still satisfy the theoretical requirement (but it may omit many other potential candidates). Therefore, our practical version algorithm should be regarded as an inaccurate approximation to the theoretical algorithm. Moreover, it also plays a role as regularization to enable the convergence of $\mathcal{F}_t$ since we only consider regular ones in the neighborhood of $\hat{f}_t$. 
\subsection{Results}
\begin{wrapfigure}{r}{0.4\textwidth}
\centering
\label{fig:res}
\vspace{-20pt}
\includegraphics[width=0.39 \textwidth]{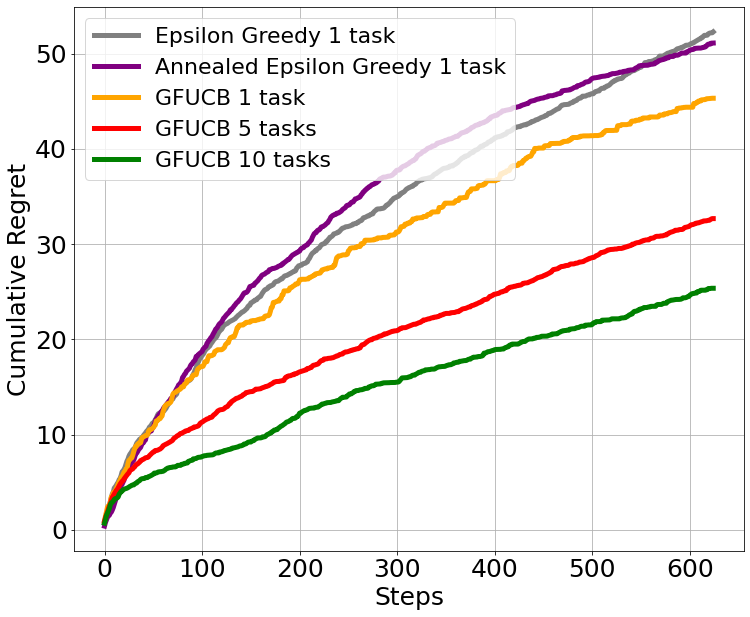}
\vspace{-10pt}
\caption{Cumulative regret over steps for $M=1,5,10$.}
\end{wrapfigure}

We test the performance of our algorithm against a naive eps-greedy baseline that solves each task independently by training \textit{the same} CNN value prediction module. We show our results with number of tasks $M = 1, 5, 10$ in \hyperref[fig:res]{Figure 1}. Firstly, we randomly generate 10 different digit-value mapping functions $\sigma_i(\cdot),i=1,\hdots,10$. The total $10$ tasks are divided into $10/M$ groups; each group forms a $M$-task problem and is solved by an individual copy of some algorithm. At each step $t$, the cumulative regret from all $10$ tasks is averaged to estimate the method's performance. Our result in \hyperref[fig:res]{Figure 1} verified that the multitask training does accelerate learning, which empirically validates our theoretical analysis. The multitask training utilizes the samples from all $M$ tasks to jointly learn a good representation $\phi$, which significantly accelerates the learning procedure of the CNN backbone. Also, the improvement in GFUCB algorithm's performance with $M=1$ validates the effect of our finetune procedure for getting a bonus. Detailed dissection and discussion are left in appendix.

%

\section{Conclusion} In this work, we extend the analysis of the benefit of multitask representation learning from linear representation class to general function class. We propose a straightforward algorithm that can utilize samples from all the tasks to jointly train a representation function, which is demonstrated theoretically and empirically to accelerate the sample efficiency and outperform naively single-task learning. Also, we extend the analysis to the MDP setting and show that the benefit of multitask representation learning is similar. Furthermore, our experimental result reveals that our proposed algorithm is also effective in practice even for highly non-linear neural network representations.

\begin{ack}
This work is supported in part by the National Science and Technology Major Project of the Ministry of Science and Technology of China under Grants 2018AAA0101604, the National Natural Science Foundation of China under Grants 62022048 and the State Key Lab of Autonomous Intelligent Unmanned Systems.
\end{ack}

\bibliography{main}
\bibliographystyle{plain}

\newpage
\appendix
\section{Bandit Regret Bound Analysis}
\subsection{Algorithm Procedure}
At each round $s\in[t]$ , after performing a list of actions $\{A_{s,i}\}_{i=1}^M$ with respect to corresponding context vectors $\{ C_{s,i} \}_{i=1}^M$, the agent receives a list of rewards $y_{s,i}$ associated with input $\bsyb{x}_{s,i}=(C_{s,i}, A_{s,i})$ for $i\in[M]$. Note that we will use $f(C_t,A_t)$ or $f(\bsyb{x}_t)$ where $\bsyb{x}_t=(C_t, A_t)$ in different contexts. The algorithm first solves the following regression problem to obtain the empirical minimizer function $\hat{f}_{t} (\cdot) = \hat{\phi}_t(\cdot)^{\top} \widehat{\bsyb{W}}_{t}$ based on samples collected. 
\begin{align*}
    \hat{\phi}_t, \widehat{\bsyb{W}}_{t} =& \mathop{\mathrm{arg min}}_{\phi \in \Phi, \bsyb{W}=[\bsyb{w}_{1,\hdots,M}]} \sum_{i=1}^M \left\| \bsyb{y}_{t-1,i} - \phi(\bsyb{X}_{t-1,i})^{\top}\bsyb{w}_i \right\|_2^2 \\
     s.t.&\quad | \phi(\bsyb{x})^{\top} \bsyb{w}_i | \leq 1,\quad  \forall i \in [M], \bsyb{x}\in \mcal{C}\times \mcal{A}.
\end{align*}
Here, $\bsyb{X}_{t-1,i} = [\bsyb{x}_{1,i},\bsyb{x}_{2,i},\hdots,\bsyb{x}_{t-1,i}]$ is the selected context-action pair for task $i$ in the first $t-1$ rounds, and $\bsyb{y}_{t-1,i} = [R_{1,i}, R_{2,i}, \hdots, R_{t-1,i}]^{\top} \in \mathbb{R}^{t-1}$ stacks all the received reward into a vector accordingly. We use $\phi(\bsyb{X})$ to compactly represent feeding each column $\bsyb{x}_i$ of $\bsyb{X}$ into $\phi(\cdot)$ and get concatenated output as $[\phi(\bsyb{x}_1), \phi(\bsyb{x}_2), \hdots, \phi(\bsyb{x}_{t-1})]$.

After obtaining the best empirical estimator function $\hat{f}_{t}^{(i)} (\cdot) = \hat{\phi}_t(\cdot)^{\top} \hat{\bsyb{w}}_{t,i}$ at round $t \in [T]$ for each $i\in[M]$, we maintain a function confidence set $\mcal{F}_t \subseteq \mcal{F}^{\otimes M}$ for representation function and parameters. 
\begin{align*}
\mathcal{F}_{t} \stackrel{\text { def }}{=} \Bigg\{ f \in \mcal{F}^{\otimes M} :  \left\| \hat{f}_{t} - f \right\|^{2}_{2,E_t} \leq \beta_t, | f^{(i)}(\bsyb{x}) | \leq 1, \forall \bsyb{x}\in \mcal{C}\times\mcal{A},i\in[M] \Bigg\} \tag{$*$}
\end{align*}
Here we abuse the notation of $\mcal{F}^{\otimes M}$ as $\mcal{F}^{\otimes M}=\left\{f=\left( f^{(1)}, \hdots, f^{(M)} \right): f^i(\cdot) = \phi(\cdot)^{\top} \bsyb{w}_i \in \mcal{F} \right\}$ to denote the M-head prediction version of $\mcal{F}$, parametrized by a shared representation function $\phi(\cdot)$ and a weight matrix $\bsyb{W}=[\bsyb{w}_1,\hdots,\bsyb{w}_M] \in \mathbb{R}^{k\times M} $. We use $f^{(i)}$ to denote the $i_{th}$ head of function $f$. For the sake of simplicity, we use 
$$\left\| \hat{f}_{t} - f \right\|^{2}_{2,E_t} = \sum_{i=1}^M \sum_{s=1}^{t-1} \left( \hat{f}_t^{(i)} (\bsyb{x}_{s,i}) - f^{(i)}(\bsyb{x}_{s,i}) \right)^2$$
to denote the empirical 2-norm of function $\hat{f}_t - f = \left( \hat{f}_t^{(1)}-f^{(1)}, \hdots, \hat{f}_t^{(M)}-f^{(M)} \right)$. 
Another important hyperparameter for our algorithm is the confidence set width term $\beta_t$, which is a function of representation function class $\Phi$, probability $\delta$ and discretization scale parameter $\alpha$. 
\begin{align*}
    \beta_t(\Phi, \alpha, \delta) = 12 Mk + 12\log \left(\mcal{N}(\Phi, \alpha, \|\cdot\|_{\infty}) / \delta\right) + 8 \alpha \sqrt{ Mtk (Mt +\log(2Mt^2 / \delta))}
\end{align*} 

here $\mcal{N}(\mcal{F}, \alpha, \|\cdot\|_{\infty})$ is the $\alpha$-covering number of function class $\Phi$ in the sup-norm $\|\phi \|_{\infty} = \max_{\bsyb{x} \in\mcal{S}\times \mcal{A}} \| \phi(\bsyb{x})\|_2$ (see detailed definition in Lemma 1) and $\alpha$ can be set to be some small scale number, like $\frac{1}{kMT}$. 
\subsection{Main Proof sketch}
In this section we will give a theoretical guarantee for the performance of our algorithm. Before diving into details, we first explain the overall idea and structure of our proof. First, we decompose the regret into the summation of confidence set width at different rounds plus a small term which accounts for the possibility that confidence function set $\mcal{F}_t$ fails to contain ground truth function $f_{\theta}$.

\textbf{Lemma 0.} \textit{Fix any sequence of confidence set $\{\mcal{F}_t, t\in \mathbb{N} \}$ which is measurable with respect to history $\mcal{H}_t$, denote the induced policy by Algorithm 1 as $\pi=\{\pi_i\}_{i=1}^M$ where each $\pi_i:\mcal{C}\mapsto \mcal{A},i\in[M]$ is for task $i$, then for any $T\in \mathbb{N}$ we have}

$$ \operatorname{Regret}(T) := \sum_{i=1}^M \sum_{t=1}^T f_{\theta}^{(i)} \left(\bsyb{x}_{t,i}^{\star}\right) - f_{\theta}^{(i)} (\bsyb{x}_{t,i}) \leq \sum_{t=1}^T \left[ w_{\mcal{F}_t} (\bsyb{X}_{t}) + C\cdot \mathbb{I}(f_{\theta} \not\in \mcal{F}_t) \right] $$

where $\bsyb{x}_{t,i} = (C_{t,i}, \pi_i(C_{t,i}))$ is the context-action pair that actually happened. $A_{t,i}^{\star} = \arg\max_{A} f_{\theta}^{(i)}(C_{t,i},A)$ is the optimal action for each task $i\in[M]$ at round $t\in [T]$, and $\bsyb{x}_{t,i}^{\star}=(C_{t,i}, A_{t,i}^{\star})$ is the corresponding optimal context-action pair, $C$ is a universal large enough constant. We use $\bsyb{X}_t=[\bsyb{x}_{t,1},\hdots, \bsyb{x}_{t,M}]$ to stack $\bsyb{x}_{t,i}$ into a matrix, similar for $\bsyb{X}_t^{\star} = [\bsyb{x}_{t,1}^{\star},\hdots, \bsyb{x}^{\star}_{t,M}]$. The confidence set width $w_{\mcal{F}_t} (\bsyb{X}_{t})$ is defined by 

$$w_{\mcal{F}_t} (\bsyb{X}_{t}) := \sup_{\overline{f},\underline{f}\in \mcal{F}_t} \sum_{i=1}^M \left[\  \overline{f}^{(i)}(\bsyb{x}_{t,i}) - \underline{f}^{(i)}(\bsyb{x}_{t,i}) \ \right]. $$

Essentially, it measures the largest total difference of value estimation among all the functions in $f\in \mcal{F}_t$ for the fixed inputs $\bsyb{x}_{t,i}$ where $i\in[M]$. Apart from the constant term accounting for the case that $\mcal{F}_t$ fails to contain $f_{\theta}$, which we will prove happen with small probability, this regret is then bounded by the sum of width over time step $t$. 

Next, we will show that our construction of confidence set $\mcal{F}_t$ makes all of them contain real value function with high probability. 

\textbf{Lemma 1.} \textit{For all $\delta\in(0,1)$ and $\alpha > 0$, if $\mcal{F}_t$ is defined by $\mcal{F}_t = \{ f\in \mcal{F}^{\otimes M}: \| f - \hat{f} \|_{2,E_t} \leq \sqrt{\beta_t(\Phi,\delta,\alpha)} \}$ for all $t\in \mathbb{N}$, where $\hat{f}$ is the solution to the empirical error minimization. Denote the ground truth value function as $f_{\theta} (\cdot)$, then we have}
$$ \mathbb{P}\left(f_{\theta} \in \bigcap_{t=1}^{T} \mcal{F}_t \right) \geq 1 - 2 \delta.$$

After that, we prove that

\textbf{Lemma 2.} $$ \sum_{t=1}^T \mathbb{I} \left( w_{\mcal{F}_t} (\bsyb{X}_{t}) > \epsilon \right) \leq \left(\frac{4 M \beta_T}{\epsilon^2} + 1 \right) \operatorname{dim}_{E}(\mcal{F}, \epsilon) $$

Then plug it into lemma 0, we get our main result for the regret bound as 

\begin{align}
    \operatorname{Reg}(\pi, T) \leq \frac{1}{T} + \min\left\{ \operatorname{dim}_E(\mcal{F}, \alpha_T), T \right\} + 4\sqrt{M \operatorname{dim}_E(\mcal{F}, \alpha_T) \beta_{T} T}
\end{align}

Usually $\alpha_T$ is set to be a small number like $\frac{1}{kMT}$, or the minimizer for $\beta_T(\Phi, \alpha, \delta)$. We know that $\operatorname{dim}_E(\mcal{F}, \alpha_T)$ is a poly-logarithmic function of $T$, which means the final regret bound is dominant by term $\sqrt{ M \operatorname{dim}_E(\mcal{F}, \alpha_T) \beta_{T} T}$ when $T \to \infty$. This further becomes
\begin{align}
    \sqrt{M T \left( Mk + \log \left(\mcal{N}(\Phi, (kMT)^{-1}, \|\cdot\|_{\infty})\right) \right) \operatorname{dim}_E(\mcal{F}, (kMT)^{-1})  }
\end{align}
For example, if $\Phi$ is specialized as linear function class parametrized by matrix $\bsyb{\Theta}\in \mathbb{R}^{d\times k}$, then $\log \left(\mcal{N}(\Phi, (kMT)^{-1}, \|\cdot\|_{\infty})\right) = O(kd\log(kMT))$ and $\operatorname{dim}_E(\mcal{F}, (kMT)^{-1}) = O(d\log (kMT))$, hence the regret bound becomes 
$$O(\sqrt{MT(Mk+kd)d} \log(k M T)) = \tilde{O}(M\sqrt{kdT} + d\sqrt{MkT})$$
which reduces to result in \cite{hu2021near} by a poly-logarithm factor.

\subsection{Detailed Proof}
\textit{Proof of Lemma 0.} Define the upper and lower bounds $U_{t} (\bsyb{X}_{t}) = \sup \left\{ \sum_{i=1}^M f^{(i)} (\bsyb{x}_{t,i}) \ :\ f\in \mcal{F}_t \right\}$ and $L_{t} (\bsyb{X}_{t}) = \inf \left\{ \sum_{i=1}^M f^{(i)} (\bsyb{x}_{t,i})\ :\ f\in \mcal{F}_t \right\}$.

If $f_{\theta} \not\in \mcal{F}_t$, then the error will be bounded by a large constant $C$ since all $f(\bsyb{x})$ is constant bounded. Otherwise $f_{\theta} \in \mcal{F}_t$, we have 
$$ L_t(\bsyb{X}_t) \leq \sum_{i=1}^M f_{\theta}^{(i)} (\bsyb{x}_{t,i}) \leq U_t(\bsyb{X}_t) $$
$$ \sum_{i=1}^M f_{\theta}^{(i)} (\bsyb{x}^{\star}_{t,i}) \leq U_t(\bsyb{X}^{\star}_t) $$

where $\bsyb{X}_t$ and $\bsyb{X}^{\star}_t$ is defined in lemma 0. Also, by the optimality of $\bsyb{X}_{t}$ with respect to $\mcal{F}_t$, we know $ U_{t} (\bsyb{X}^{\star}_{t}) \leq  U_{t} (\bsyb{X}_{t})$, therefore
\begin{align*}
        \sum_{i=1}^M \left[ f_{\theta}^{(i)} (\bsyb{x}^{\star}_{t,i}) - f_{\theta}^{(i)} (\bsyb{x}_{t,i}) \right]
    \leq& C \cdot  \mathbb{I}(f_{\theta} \not\in \mcal{F}_t) + \left[ U_{t} (\bsyb{X}^{\star}_{t}) - L_{t} (\bsyb{X}_{t}) \right] \\
    =& C \cdot \mathbb{I}(f_{\theta} \not\in \mcal{F}_t) +  \sum_{i=1}^M \left[ U_{t} (\bsyb{X}^{\star}_{t}) - U_{t} (\bsyb{X}_{t}) + U_{t} (\bsyb{X}_{t}) - L_{t} (\bsyb{X}_{t}) \right] \\
    \leq& C \cdot \mathbb{I}(f_{\theta} \not\in \mcal{F}_t) +  \sum_{i=1}^M \left[ U_{t} (\bsyb{X}_{t}) - L_{t} (\bsyb{X}_{t}) \right]  \\
    =& C \cdot \mathbb{I}(f_{\theta} \not\in \mcal{F}_t) +  w_{\mcal{F}_t}(\bsyb{X}_t)
\end{align*}
Take summation over $t \in [T]$ and complete the proof.
\qed \\

\textbf{Lemma 1.} \textit{For all $\delta\in(0,1)$ and $\alpha > 0$, if $\mcal{F}_t$ is defined by $\mcal{F}_t = \left\{ f\in \mcal{F}^{\otimes M}: \| f - \hat{f} \|_{2,E_t} \leq \sqrt{\beta_t(\Phi,\delta,\alpha)} \right\}$ for all $t\in \mathbb{N}$, where $\hat{f}$ is the solution to the empirical error minimization. Denote the ground truth value function as $f_{\theta}$, then we have}

$$ \mathbb{P}\left(f_{\theta} \in \bigcap_{t=1}^{T} \mcal{F}_t \right) \geq 1 - 2 \delta.$$

\textit{Proof of Lemma 1.} Denote $L_{2,t}(f) = \sum_{i=1}^M \sum_{s=1}^t |f^{(i)} (\bsyb{x}_{s,i}) - y_{s,i}|^2$ and $\tilde{f}_t = \hat{f}_t-f_{\theta}$, we have
\begin{align}
    L_{2,t}(\hat{f}) - L_{2,t}(f_{\theta}) =& \sum_{i=1}^M \sum_{s=1}^t \left| \hat{f}^{(i)}_t (\bsyb{x}_{s,i}) - y_{s,i} \right|^2 - \left| f_{\theta}^{(i)} (\bsyb{x}_{s,i}) - y_{s,i} \right|^2 \\
    =& \sum_{i=1}^M \sum_{s=1}^t \left| \hat{f}^{(i)}_t (\bsyb{x}_{s,i}) - f_{\theta}^{(i)} (\bsyb{x}_{s,i}) - \eta_{s,i} \right|^2 - \eta_{s,i}^2 \\
    =& \left\| \hat{f}_t - f_{\theta} \right\|_{2,E_t}^2 - \sum_{i=1}^M \sum_{s=1}^t 2\eta_{s,i} \cdot \tilde{f}^{(i)}_t (\bsyb{x}_{s,i})
\end{align}
By the optimality of $\hat{f}$, we know (5) $\leq 0$, hence
\begin{align}
    \left\| \hat{f}_t - f_{\theta} \right\|_{2,E_t}^2 \leq  \sum_{i=1}^M  2 \left\langle \bsyb{\eta}_{t,i}, \tilde{f}^{(i)}_t (\bsyb{X}_{t,i}) \right\rangle
\end{align}

here $\tilde{f}^{(i)}_t (\bsyb{X}_{t,i}) = [\tilde{f}^{(i)}_t (\bsyb{x}_{1,i}), \tilde{f}^{(i)}_t (\bsyb{x}_{2,i}), \hdots, \tilde{f}^{(i)}_t (\bsyb{x}_{t,i})]^{\top} $ and $\bsyb{\eta}_{t,i}=[\eta_{1,i}, \eta_{2,i}, \hdots, \eta_{t,i}]^{\top}$ are both in $\mathbb{R}^t$. We can represent each function $\tilde{f}_t^{(i)}(\cdot)$ in form $\tilde{f}_t^{(i)} (\cdot) = \left[\phi^{\star}(\cdot)^{\top}, \hat{\phi}_t (\cdot)^{\top} \right] \left[ \begin{matrix} \bsyb{w}^{\star}_{t,i} \\ -\hat{\bsyb{w}}_{t,i} \end{matrix} \right] = \phi^{\star}(\cdot)^{\top}\bsyb{w}^{\star}_{t,i} - \hat{\phi}_t (\cdot)^{\top}\hat{\bsyb{w}}_{t,i}$, which is exactly $f_{\theta} - \hat{f}_t$. Denote $\tilde{\phi}_t (\cdot) = \left[ \begin{matrix} \phi^{\star}(\cdot) \\ \hat{\phi}_t (\cdot) \end{matrix} \right] \in \Phi^2$ and $\tilde{\bsyb{w}}_{t,i} = \left[ \begin{matrix} \bsyb{w}^{\star}_{t,i} \\ -\hat{\bsyb{w}}_{t,i} \end{matrix} \right]\in \mathbb{R}^{2k}$, then $\tilde{f}_{t}^{(i)}(\cdot) = \tilde{\phi}_{t}(\cdot) ^{\top} \tilde{\bsyb{w}}_{t,i}$. Since the output of $\tilde{\phi}_t (\bsyb{x}_{s,i}) \in \mathbb{R}^{2k}$, we can take following decomposition for each $i\in [M]$

$$ \tilde{\phi}_t (\bsyb{X}_{t,i}) = \left[ \tilde{\phi}_t (\bsyb{x}_{s,i}) \right]_{s=1}^{t},\quad \tilde{\phi}_t (\bsyb{X}_{t,i})^{\top} = \bsyb{U}_i \bsyb{Q}_i,\quad \bsyb{U}_i \in \mcal{O}^{t\times 2k}, \bsyb{Q}_i \in \mathbb{R}^{2k \times 2k}. $$

For regret bound, we only need to care about $t \geq 2k$ by a constant regret difference, hence this decomposition is possible. Plug it into (6) and we get
\begin{align}
    \frac{1}{2} \left\| \hat{f} - f_{\theta} \right\|_{2,E_t}^2 \leq& \sum_{i=1}^M  \left\langle \bsyb{\eta}_{t,i}, \tilde{f}^{(i)}_t (\bsyb{X}_{t,i}) \right\rangle \\
    =& \sum_{i=1}^M  \bsyb{\eta}_{t,i}^{\top} \cdot \tilde{\phi}_t (\bsyb{X}_{t,i})^{\top} \tilde{\bsyb{w}}_{t,i} \\
    =& \sum_{i=1}^M  \bsyb{\eta}_{t,i}^{\top} \cdot \bsyb{U}_i \bsyb{Q}_i \tilde{\bsyb{w}}_{t,i}
\end{align}

Notice that, however, $\bsyb{U}_t$ is obtained from optimization problem, which further depends on concrete sampled noise $\bsyb{\eta}_{t,i}$, hence the concentration bound based on i.i.d. assumption cannot be applied directly. If we fix function $\tilde{f}_t = \bar{f}_t$, which induces corresponding $\bar{\phi}_t(\cdot)$ and $\bar{\phi}_t(\bsyb{X}_{t,i})=\bar{\bsyb{U}}_i(\bar{\phi}) \bar{\bsyb{Q}}_i$, $\bar{\bsyb{U}}_i(\bar{\phi})$ means $\bar{\bsyb{U}}_i$ is a function determined by $\bar{\phi}$. According to standard sub-exponential random variable concentration bound, each $\bar{\bsyb{U}}_i (\bar{\phi})$ has $2k$ independent degrees of freedom, hence we know that with probability at least $1-\delta_1$ 
\begin{align}
    \sum_{i=1}^{M} \| \bar{\bsyb{U}}_i^{\top} \bsyb{\eta}_{t,i} \|^2 \leq 2 Mk + \log (1/\delta_1)
\end{align}

Denote $\Phi^2=\{g(\bsyb{x})=[\phi_1(\bsyb{x})^{\top}, \phi_2(\bsyb{x})^{\top}]^{\top} : \phi_1, \phi_2 \in \Phi\}$, $\Phi^2_{\alpha}$ is an $\alpha$-cover of $\Phi^2$ such that for any $\phi \in \Phi^2$, there is a $\phi_{\alpha} \in \Phi^2_{\alpha}$ such that 
\begin{align}
    \max_{\bsyb{x}\in \mcal{C} \times \mcal{A}} \| \phi(\bsyb{x}) - \phi_{\alpha}(\bsyb{x}) \|_2 \leq \alpha. 
\end{align} 
 For $\tilde{\phi}$, find a closest $\bar{\phi} \in \Phi^2_{\alpha}$ from $\alpha$-cover net to satisfy the requirement above, then denote $\bar{f}_{t}^{(i)}(\cdot) = \bar{\phi}(\cdot)^{\top} \tilde{\bsyb{w}}_{t,i}$. By union bound, we know that with probability at least $1-|\Phi_{\alpha}^2| \delta_1$, for any $\bar{\phi} \in \Phi_{\alpha}^2$, the induced $\bar{\bsyb{U}}_{i}(\bar{\phi})$ satisfy inequality (10), therefore
\begin{align}
    \frac{1}{2}\left\| \hat{f}_t - f_{\theta} \right\|_{2,E_t}^2 \leq& \sum_{i=1}^M \left\langle \bsyb{\eta}_{t,i}, \tilde{f}_{t}^{(i)}( \bsyb{X}_{t,i} ) \right\rangle \\
    =& \sum_{i=1}^M \bsyb{\eta}_{t,i}^{\top} \cdot \bsyb{U}_i \bsyb{Q}_i \tilde{\bsyb{w}}_{t,i} =\sum_{i=1}^M \bsyb{\eta}_{t,i}^{\top} \cdot (\bsyb{U}_i - \bar{\bsyb{U}}_i + \bar{\bsyb{U}}_i) \bsyb{Q}_i \tilde{\bsyb{w}}_{t,i} \\
    =&\sum_{i=1}^M \bsyb{\eta}_{t,i}^{\top} \cdot \bar{\bsyb{U}}_i \bsyb{Q}_i \tilde{\bsyb{w}}_{t,i} + \sum_{i=1}^M\bsyb{\eta}_{t,i}^{\top} \cdot (\bsyb{U}_i - \bar{\bsyb{U}}_i) \bsyb{Q}_i \tilde{\bsyb{w}}_{t,i}\\
    \leq& \sqrt{\sum_{i=1}^M \left\| \bar{\bsyb{U}}_i^{\top} \bsyb{\eta}_{t,i} \right\|^2} \cdot \sqrt{\sum_{i=1}^M \left\| \bsyb{Q}_i \tilde{\bsyb{w}}_{t,i} \right\|^2} + \sum_{i=1}^M \left\langle \bsyb{\eta}_{t,i}, \tilde{f}_t - \bar{f}_t \right\rangle \\
    \leq& \sqrt{\sum_{i=1}^M \left\| \bar{\bsyb{U}}_i^{\top} \bsyb{\eta}_{t,i} \right\|^2} \cdot \sqrt{\sum_{i=1}^M \left\|  \bsyb{U}_i \bsyb{Q}_i \tilde{\bsyb{w}}_{t,i} \right\|^2} + \sum_{i=1}^M \left\langle \bsyb{\eta}_{t,i}, \tilde{f}_t - \bar{f}_t \right\rangle \\
    =& \sqrt{\sum_{i=1}^M \left\| \bar{\bsyb{U}}_i^{\top} \bsyb{\eta}_{t,i} \right\|^2} \cdot \left\| \tilde{f} \right\|_{2, E_t} + \sum_{i=1}^M \left\langle \bsyb{\eta}_{t,i}, \tilde{f}_t - \bar{f}_t \right\rangle \\
  \leq& \sqrt{2Mk + \log (1/\delta_1)} \cdot \left\| \tilde{f} \right\|_{2, E_t} + \sqrt{\sum_{i=1}^M \| \bsyb{\eta}_{t,i} \|^2} \cdot  \left\| \tilde{f}_t - \bar{f}_t \right\|_{2,E_t}
\end{align}
The first term of (18) comes from (10), and the second term is from Cauchy inequality. We assign $\delta_t = \frac{\delta_2}{T}$ failure probability for event 
$$ \omega_t: \sum_{i=1}^M \| \bsyb{\eta}_{t,i} \|^2 \geq Mt + \log(2Mt/\delta_t). $$
By union bound, we have
\begin{align}
    \mathbb{P}\left( \exists t\in [T]: \sum_{i=1}^M \| \bsyb{\eta}_{t,i} \|^2 \geq Mt + \log(2M t^2/\delta_2) \right) \leq \sum_{t=1}^{T} \delta_t \leq \delta_2.
\end{align}
Next we will give a bound for $\| \tilde{f}_t - \bar{f}_t \|_{2,E_t}$.
\begin{align}
    \left\| \tilde{f}_t - \bar{f}_t \right\|_{2,E_t}^2 =& \sum_{i=1}^M \sum_{s=1}^t \left| \tilde{\phi}_t(\bsyb{x}_{s,i})^{\top} \tilde{\bsyb{w}}_{s,i} - \bar{\phi}_t(\bsyb{x}_{s,i})^{\top} \tilde{\bsyb{w}}_{s,i} \right|^2 \\
    =& \sum_{i=1}^M \sum_{s=1}^t \left| ( \tilde{\phi}_t(\bsyb{x}_{s,i}) - \bar{\phi}_t(\bsyb{x}_{s,i}) )^{\top}\tilde{\bsyb{w}}_{s,i} \right|^2 \\
    \leq& \sum_{i=1}^M \sum_{s=1}^t \left\| \tilde{\phi}_t(\bsyb{x}_{s,i}) - \bar{\phi}_t(\bsyb{x}_{s,i})\right\|_2^2 \cdot \left\| \tilde{\bsyb{w}}_{s,i} \right\|_2^2 
\end{align}
According to our assumption, we know $\left\| \tilde{\bsyb{w}}_{s,i} \right\|^2 \leq 2\|\bsyb{w}_{s,i}\|^2 + 2\|\hat{\bsyb{w}}_{s,i}\|^2 \leq 4k $, from (11) we know $ \left\| \tilde{\phi}_t(\bsyb{x}_{s,i}) - \bar{\phi}_t(\bsyb{x}_{s,i})\right\|_2 \leq \alpha$, hence
\begin{align}
    \left\| \tilde{f}_t - \bar{f}_t \right\|_{2,E_t}^2 \leq& 4Mtk \alpha^2
\end{align}
Plug (19) and (23) back into (18), we know with probability at least $1-\delta_2 - |\Phi_{\alpha}^2|\delta_1$, for any $t\in\mathbb{N}$
\begin{align}
    \frac{1}{2}\left\| \tilde{f}_t \right\|_{2,E_t}^2 \leq&  \sqrt{2Mk + \log (1/\delta_1)} \cdot \left\| \tilde{f}_t \right\|_{2, E_t} + \sqrt{Mt + \log(2 M t^2 / \delta_2)} \cdot \sqrt{4 M t k \alpha^2}
\end{align}
Some simple algebraic transform gives
\begin{align}
     \left\| \hat{f}_t - f_{\theta} \right\|_{2,E_t}^2 =& \left\| \tilde{f}_t \right\|_{2,E_t}^2 \leq 6(2Mk + \log (1/\delta_1)) + 8 \alpha \sqrt{Mtk (Mt + \log(2 M t^2 / \delta_2))}
\end{align}
Let $\delta_1 = \delta / |\Phi_{\alpha}^2|, \delta_2 = \delta$, and notice $\log |\Phi_{\alpha}^2| \leq 2 \log \left(\mcal{N}(\Phi, \alpha, \|\cdot\|_{\infty})\right)$, we conclude that with probability at least $1-2\delta$, for every $t\in\mathbb{N}$
\begin{align}
     \left\| \hat{f}_t - f_{\theta} \right\|_{2,E_t}^2 \leq 12 Mk + 12\log \left(\mcal{N}(\Phi, \alpha, \|\cdot\|_{\infty}) / \delta\right) + 8 \alpha \sqrt{Mtk (Mt + \log(2Mt^2 / \delta))}
\end{align}
where the right handside is exactly our defined $\beta_t(\Phi, \alpha, \delta)$, hence our conclusion holds.
\qed
\\

\textbf{Lemma 2.} \textit{If $(\beta_t \geq 0 \mid t \in \mathbb{N})$ is a nondecreasing sequence and $\mcal{F}_t := \left\{ f\in \mcal{F}^{\otimes M}: \|f - \hat{f}_{t}^{LS}\|_{2,E_t} \leq \sqrt{\beta_t} \right\}$. Also, denote $\mcal{F}=\mcal{L} \circ \Phi : \mcal{C}\times \mcal{A}\mapsto [0,1]$, we have}
$$ \sum_{t=1}^T \mathbb{I}\left( w_{\mcal{F}_t}(\bsyb{X}_{t}) > \epsilon \right) \leq \left( \frac{4 M \beta_T}{\epsilon^2}+1 \right) \operatorname{dim}_E(\mcal{F}, \epsilon) $$
\textit{Proof.} The main structure of this proof is similar to proposition 3, section C in Eluder dimension's paper, and we will only point out the subtle details that makes the difference. We will show that if $w_{\mcal{F}_t}(\bsyb{X}_t) > \epsilon$ , then $\bsyb{X}_t$ is $\epsilon$-dependent on fewer than $4M\beta_T/\epsilon^2$ disjoint subsequences of $(\bsyb{X}_{1},\hdots, \bsyb{X}_{t-1})$. Note that if $w_{\mcal{F}_t}(\bsyb{X}_t) > \epsilon$, there are $\overline{f},\underline{f} \in \mcal{F}_t$ such that $\sum_{i=1}^M \overline{f}^{(i)}(\bsyb{x}_{t,i})-\underline{f}^{(i)}(\bsyb{x}_{t,i}) > \epsilon$. By definition, if $\bsyb{X}_t$ is $\epsilon$-dependent on a subsequence $(\bsyb{X}_{t_1}, \bsyb{X}_{t_2}, \hdots, \bsyb{X}_{t_k})$ of $(\bsyb{X}_{1},\hdots, \bsyb{X}_{t-1})$, then we know

$$ \sum_{j=1}^k \left( \sum_{i=1}^M \overline{f}^{(i)}(\bsyb{x}_{t_j,i})-\underline{f}^{(i)}(\bsyb{x}_{t_j,i}) \right)^2 > \epsilon^2 $$

It follows that, if $\bsyb{X}_t$ is $\epsilon$-dependent on $K$ disjoint subsequences of $(\bsyb{X}_1, \hdots, \bsyb{X}_{t-1})$, then 
\begin{align}
    \| \overline{f} - \underline{f} \|_{2,E_t}^2 =&  \sum_{s=1}^t \sum_{i=1}^M \left( \overline{f}^{(i)}(\bsyb{x}_{s,i}) - \underline{f}^{(i)}(\bsyb{x}_{s,i}) \right)^2 \\
    \geq& \frac{1}{M} \sum_{s=1}^t \left(  \sum_{i=1}^M \overline{f}^{(i)}(\bsyb{x}_{s,i}) - \underline{f}^{(i)}(\bsyb{x}_{s,i}) \right)^2 \tag{Cauchy Inequality} \\
    >& \frac{K \epsilon^2}{M}
\end{align} 

By triangle inequality we have
\begin{align}
    \| \overline{f} - \underline{f} \|_{2,E_t} \leq \| \overline{f} - \hat{f}^{LS}_t \|_{2,E_t} + \| \hat{f}^{LS}_t - \underline{f} \|_{2,E_t} \leq 2\sqrt{\beta_t} \leq 2\sqrt{\beta_T}
\end{align}
and it follows that $K < 4 M \beta_T / \epsilon^2 $.

Notice that essentially we are analyzing scalar output function $g(\bsyb{X}_t) = \sum_{i=1}^M f^{(i)}(\bsyb{x}_{t,i})$ where $f\in\mcal{F}^{\otimes M}$. Hence if we denote any $f\in\mcal{F}^{\otimes M}$ as $f(\cdot) = \phi(\cdot)^{\top} \bsyb{\Theta}$, then $g(\cdot) = \phi(\cdot)^{\top} \bsyb{w}\in \mcal{F}, \bsyb{w}=\bsyb{\Theta} \cdot \bsyb{1}$. Hence from original eluder dimension paper we know in any action sequence $(\bsyb{X}_1, \hdots, \bsyb{X}_{\tau})$, there must exist some element $\bsyb{X}_j$ that is $\epsilon$-dependent on at least $\tau / d - 1$ disjoint subsequences of $(\bsyb{X}_1, \hdots, \bsyb{X}_{\tau})$, where $d:= \operatorname{dim}_E (\mcal{F}, \epsilon)$. Finally we select $\bsyb{X}_1,\hdots, \bsyb{X}_{\tau}$ as those actions that $w_{\mcal{F}_t} > \epsilon$, combine these two facts above and get $\tau / d - 1 \leq 4 M \beta_T / \epsilon^2$. Hence $\tau \leq (4M\beta_T/\epsilon^2 + 1)d$, which is our desired conclusion.

\section{Linear MDP Regret Analysis}
Apart from the notations section 3, we add more symbols for the regret analysis. We use $Q[f]$ or $Q[\phi\circ \bsyb{\theta}]$ to denote the Q-value function parametrized by function $f$ as $Q[f](s,a) = f(s,a)$ or $Q[\phi\circ \bsyb{\theta}](s,a) = \phi(s,a)^{\top} \bsyb{\theta}$ (similar for $V[f]$ as state's value estimation function). Also, based on assumption 2.1, for any $ \left\{Q_{h+1}^{(i)} \right\}_{i=1}^M$, there always exists $\dot{f}_{h} \left[ Q_{h+1} \right] \in \mcal{F}^{\otimes M}$ such that
\begin{align}
    \Delta_{h}^{(i)} \left( Q_{h+1}^{(i)} \right) (s,a) = \mcal{T}_{h}^{i} \left( Q_{h+1}^{(i}) \right) (s,a) - \dot{f}_{h}^{(i)}(s,a)
\end{align}
where the approximation error $\left\| \Delta_{h}^{(i)} \left( Q_{h+1}^{(i)} \right) \right\| \leq \mcal{I}$ for $\forall \ i \in [M]$. Here $\dot{f}_{h}[Q_{h+1}]$ indicates that function $\dot{f}_h$ has dependence on Q-value function $Q_{h+1}$ on next level $h+1$. In following analysis, we will use different annotations for different function approximation as below
\begin{itemize}
    \item $f^{(i)*}_h(\cdot, \cdot) = \phi^*(\cdot, \cdot)^{\top} \bsyb{\theta}_{h}^{(i)*}$ is the ``best'' Q-value function approximation in $\mcal{Q}_h$ for task $i$ at level $h$.
    \item $\hat{f}^{(i)}_h (\cdot, \cdot) = \hat{\phi}(\cdot, \cdot)^{\top} \hat{\bsyb{\theta}}_{i}$ is the empirical least-square minimizer solution for task $i$ at level $h$.
    \item $\dot{f}^{(i)}_h (\cdot, \cdot) = \dot{\phi}(\cdot, \cdot)^{\top} \dot{\bsyb{\theta}}_{i}$ is the value approximation function $\mcal{T}_{h}^{(i)} Q_{h+1}^{(i)}$ induced by $Q_{h+1}^{(i)}$ for task $i$ at level $h$.
    \item $\tilde{f}^{(i)}_h (\cdot, \cdot) = \tilde{\phi}(\cdot, \cdot)^{\top} \tilde{\bsyb{\theta}}_{i}$ is the optimism Q-value approximation function for task $i$ at level $h$.
    \item $\bar{f}^{(i)}_h (\cdot, \cdot) = \bar{\phi}(\cdot, \cdot)^{\top} \bar{\bsyb{\theta}}_{i}$ is the nearest neighbor in covering set for task $i$ at level $h$.
\end{itemize}

\subsection{Main Proof sketch}
The overall structure is similar to bandits, the main difference here is that we need to take care of the transition dynamics. 

Firstly, we decompose the total regret into following terms
\begin{align}
    \operatorname{Reg}(T) =& \sum_{t=1}^T \sum_{i=1}^M \left( V_1^{(i)\star} - V_{1}^{\pi_t^i} \right) \left( s_{1,t}^{(i)} \right) \\
    =& \sum_{t=1}^T \sum_{i=1}^M  \left( V_1^{(i)\star} - V_{1}^{(i)} \left[ \tilde{f}_{1,t}^{(i)} \right] \right) \left(s_{1,t}^{(i)} \right) + \sum_{t=1}^T \sum_{i=1}^M  \left( V_{1}^{(i)} \left[ \tilde{f}_{1,t}^{(i)} \right] - V_{1}^{\pi_t^i} \right) \left( s_{1,t}^{(i)} \right) \\
    \leq& \sum_{t=1}^T \sum_{i=1}^M  \left( V_{1}^{(i)} \left[ \tilde{f}_{1,t}^{(i)} \right] - V_{1}^{\pi_t^i} \right) \left( s_{1,t}^{(i)} \right) + MHT \mcal{I}.
\end{align}
The inequality is because according to lemma 3, we have at each episode $t\in[T]$
\begin{align*}
      \sum_{i=1}^M  \left( V_1^{i\star} - V_{1}^{(i)} \left[ \tilde{f}_{1,t}^{(i)} \right] \right) \left( s_{1,t}^{(i)} \right) \leq& MH \mcal{I} \\
      \Longrightarrow \sum_{t=1}^T \sum_{i=1}^M  \left( V_1^{i\star} - V_{1}^{(i)} \left[ \tilde{f}_{1,t}^{(i)} \right] \right) \left( s_{1,t}^{(i)} \right) \leq& MHT \mcal{I}.
\end{align*}
Denote $a_{h,t}^{(i)} = \pi_{t}^i \left( s_{ht}^{(i)} \right)$, $Q_h^{(i)} [\tilde{f}_{h,t}^{(i)} ] = \tilde{Q}_{h,t}^{(i)}$ and $V_h^{(i)} [\tilde{f}_{h,t}^{(i)} ] = \tilde{V}_{h,t}^{(i)}$ for short. We have for any $t\in[T], h\in[H]$
\begin{align}
    \sum_{i=1}^M  \left( \tilde{V}_{h,t}^{(i)} - V_{h,t}^{\pi_t^i} \right) \left( s_{h,t}^{(i)} \right) =& \sum_{i=1}^M  \left( \tilde{Q}_{h,t}^{(i)} - Q_{h,t}^{\pi_t^i} \right) \left( s_{h,t}^{(i)} ,\  a_{h,t}^{(i)} \right) \\
    =& \sum_{i=1}^M  \left( \tilde{Q}_{h,t}^{(i)} - \mcal{T}_{h}^{(i)} \tilde{Q}_{h+1,t}^{(i)} \right) \left( s_{1,t}^{(i)} ,\  a_{h,t}^{(i)} \right) + \sum_{i=1}^M  \left(  \mcal{T}_{h}^{(i)} \tilde{Q}_{h+1,t}^{(i)} - Q_{h,t}^{\pi_t^i} \right) \left( s_{h,t}^{(i)} ,\  a_{h,t}^{(i)} \right)
\end{align}

Since the failure event $\bigcup_{t=1}^T \bigcup_{h=1}^H E_{ht}$ only happens with probability $\delta$ according to lemma 6, and the addition of regret when it happens is constant bounded, we will simply assume that it does not happen. Then applying lemma 5, we have 
\begin{align}
    \sum_{i=1}^M  \left( \tilde{Q}_{h,t}^{(i)} - \mcal{T}_{h}^{(i)} \tilde{Q}_{h+1,t}^{(i)} \right) \left( s_{h,t}^{(i)} ,\  a_{h,t}^{(i)} \right) \leq M \mcal{I} + 2 w_{\mcal{F}_{h,t}} \left( \bsyb{x}_{h,t} \right).
\end{align}
 where $\bsyb{x}_{h,t} = \left[(s_{h,t}^{(1)}, a_{h,t}^{(1)}), \hdots, (s_{h,t}^{(M)}, a_{h,t}^{(M)}) \right]$ denotes the stacked input for all state-action pair at level $h$, episode $t$.

Next, we expand the second summation in (35) and have 
\begin{align}
    \sum_{i=1}^M  \left(  \mcal{T}_{h}^{(i)} \tilde{Q}_{h+1,t}^{(i)} - Q_{h,t}^{\pi_t^i} \right) \left( s_{h,t}^{(i)} ,\  a_{h,t}^{(i)} \right) =& \sum_{i=1}^M \mathbb{E}_{s'\sim \mcal{P}_h^{(i)} \left( \cdot|s_{h,t}^{(i)}, a_{h,t}^{(i)} \right) }\left[ \left( \tilde{V}_{h+1,t}^{(i)} - V_{h+1}^{\pi_t^i} \right) (s') \right] \\
    =& \sum_{i=1}^M \left( \tilde{V}_{h+1,t}^{(i)} - V_{h+1}^{\pi_t^i} \right) \left( s_{h+1,t}^{(i)} \right) + \sum_{i=1}^M \zeta_{h,t}^{(i)}
\end{align}
where $\zeta_{h,t}^{(i)}$ is a martingale difference with respect to history $\mathcal{H}_{h,t}$ defined by
\begin{align}
\zeta_{h,t}^{(i)} \stackrel{\text { def }}{=}  \mathbb{E}_{s'\sim \mcal{P}_h^{(i)} \left( \cdot|s_{h,t}^{(i)}, a_{h,t}^{(i)} \right) }\left[ \left( \tilde{V}_{h+1,t}^{(i)} - V_{h+1}^{\pi_t^i} \right) (s') \right] - \left( \tilde{V}_{h+1,t}^{(i)} - V_{h+1}^{\pi_t^i} \right) (s')
\end{align}
According to assumption 2.2 we know that $ |\zeta_{h,t}^{(i)}| 
\leq 4$, hence by Azuma-Hoeffding's inequality, we know that with probability at least $1-\delta/2$, for any $t\in[T]$ and $i \in [M]$
\begin{align}
    \sum_{j=1}^t \zeta_{h,t}^{(i)} \leq 4 \sqrt{ 2t \log \frac{2T}{\delta}}.
\end{align}
We can then apply (38) recursively from $h=1$ to $H$, which gives
\begin{align}
    \operatorname{Reg}(T) \leq& \sum_{t=1}^T \sum_{i=1}^M \left( \tilde{V}_{1,t}^{(i)} - V_1^{\pi_t^i} \right) \left( s_{1,t}^{(i)} \right) + MHT \mcal{I} \\
    \leq& 2MHT\mcal{I} + \sum_{t=1}^T \sum_{h=1}^H 2 w_{\mcal{F}_t} (\bsyb{x}_{h,t}) + \sum_{i=1}^M \sum_{h=1}^H \sum_{t=1}^T \zeta_{h,t}^{(i)}
\end{align}
According to lemma 2 we know that
\begin{align}
     \sum_{t=1}^T w_{\mcal{F}_t} (\bsyb{x}_{h,t}) \leq \left( \frac{4 M \beta_{h,T}}{\alpha^2} +1 \right) \operatorname{dim}_{E}(\mcal{F}, \alpha )
\end{align}
where $\beta_{h,t} = \tilde{O} (Mk+\log\mcal{N}(\Phi, \alpha, \|\cdot\|_{\infty}) + MT\mcal{I}^2)$. Summarizing all inequality above and we have the final regret bound as
\begin{align}
    \operatorname{Reg}(T) =& 2 M H T \mcal{I} + \sum_{t=1}^T \sum_{h=1}^H 2 w_{\mcal{F}_t} (\bsyb{x}_{h,t}) + \sum_{i=1}^M \sum_{h=1}^H \sum_{t=1}^T \zeta_{h,t}^{(i)} \\
    =& \tilde{O} \left( MHT\mcal{I} + \tilde{O}(\sqrt{Mk+\log\mcal{N}(\Phi, \alpha, \|\cdot\|_{\infty}) + MT\mcal{I}^2}) H\sqrt{M T \dim_{E}(\mcal{F}, \alpha) } + M H \sqrt{T} \right)
\end{align}
Set $\alpha=\frac{1}{kMT}$, we have the regret bound as 
$$ \tilde{O} \left( H\sqrt{\dim_{E}(\mcal{F}, (kMT)^{-1})} \left( M \sqrt{Tk} + \sqrt{M T \log\mcal{N}(\Phi, (kMT)^{-1}, \|\cdot\|_{\infty}) } + M T \mcal{I}  \right) \right). $$

\subsection{Detailed Lemma Proof}
\textbf{Lemma 3.} \textit{Let $ V_1^{i\star} $ be the value of optimal policy and $ V_{1}^i \left[ \tilde{f}_{1,t}^{(i)} \right]$ be the optimistic value estimation defined in main proof. We have the accuracy guarantee as}
\begin{align}
    \sum_{i=1}^M  \left( V_1^{(i)\star} - V_{1}^{(i)} \left[ \tilde{f}_{1,t}^{(i)} \right] \right) \left( s_{1,t}^{(i)} \right) \leq& MH \mcal{I}.
\end{align}
\textit{Proof.} 
Recursively define the closest value approximator function $f^{*}_h = (\phi_h^*)^{\top} \bsyb{\Theta}_{h}^{*} $ at level $h$ within function class $\mcal{F}^{\otimes M}$ as
\begin{align}
    \phi_h^*, \bsyb{\Theta}_{h}^{*} \stackrel{\text { def }}{=} \mathop{\arg\min}_{\phi \in \Phi, \bsyb{\Theta}=[\bsyb{\theta}_1,\hdots,\bsyb{\theta}_M]\in \mathbb{R}^{k\times M}} \ \sup_{s,a,i} \left| \phi(s,a)^{\top} \bsyb{\theta}_h^{(i)} - \mcal{T}_h^{(i)} Q_{h+1}^{(i)} \left[ \phi_{h+1}^* \circ \bsyb{\theta}_{h+1}^{(i)*} \right](s,a) \right|
\end{align}
with $\bsyb{\theta}_{H+1}^{(i)} = \bsyb{0}$ for any $i\in[M]$ and $\bsyb{\Theta}_h^* = \left[ \bsyb{\theta}_{h}^{(1)*}, \hdots, \bsyb{\theta}_{h}^{(M)*} \right]$. By lemma 6 in \cite{zanette2020learning} we have
\begin{align}
    \sup_{(s,a)\in \mcal{S}\times\mcal{A}, i\in[M]} \left| Q_h^{(i)\star}(s,a) - \phi^*_h(s,a)^{\top} \bsyb{\theta}_{h}^{(i)*} \right| \leq (H-h+1) \mcal{I}.
\end{align}
where $Q_h^{(i)\star}$ is the optimal value function for task $i$.

Next, we will show that $f_{h}^{*}$ is a feasible solution for the optimization of $\mcal{F}_t$. This is achieved via inductive construction. For $h=H+1$ we know it holds trivially because $\tilde{f}_{H+1}^{(i)} = f_{H+1}^{(i)*} = \bsyb{0}$. Now we suppose that $\beta_{h,t}$ for $k=h+1,\hdots, H$ satisfies that we can always find  $\tilde{f}_{k}^{(i)} = f_{k}^{(i)*}$. Then from the definition of $f_h^{(i)*}$ we can always properly set $\mcal{F}_{h,t}$ (to be specified later) to let it contain
\begin{align}
    \dot{f}_h^{(i)} \left[ V_{h+1}^{(i)}\left[ f_{h+1}^{(i)*} \right] \right] = f_{h}^{(i)*}.
\end{align}

By lemma 4, we have
\begin{align}
    \left\| \hat{f}_h  \left[ V_{h+1}\left[ f_{h+1}^{*} \right] \right]  -  \dot{f}_h \left[ V_{h+1}\left[ f_{h+1}^{*} \right] \right] \right\|_{2,E_t}^2 \leq \beta_{h,t}.
\end{align}
Therefore, set $\beta_{h,t}$ as the function we set \textit{does} let $f_h^{(i)*} \in \mcal{F}_{h,t}$.

Finally, we can finish the proof from showing that
\begin{align}
    &\sum_{i=1}^M V_1^{(i)}\left[ \tilde{f}_{1,t}^{(i)} \right] \left( s_{1,t}^{(i)} \right) \\
    =& \sum_{i=1}^M \max_{a \in \mcal{A}} \tilde{f}_{1,t}^{(i)} \left( s_{1,t}^{(i)}, a \right) \\
    \geq&  \sum_{i=1}^M \max_{a \in \mcal{A}} {f}_{1,t}^{(i)*} \left( s_{1,t}^{(i)}, a \right) \tag{because $f_{1}^{(i)*} \in \mcal{F}_t$} \\
    \geq&  \sum_{i=1}^M {f}_{1,t}^{(i)*} \left( s_{1,t}^{(i)}, \pi_1^{i\star}  \left( s_{1,t}^{(i)} \right) \right) \\
    \geq& \sum_{i=1}^M Q_{1}^{(i)\star} \left( s_{1,t}^{(i)}, \pi_1^{i\star}  \left( s_{1,t}^{(i)} \right) \right) - M H \mcal{I} \tag{By (48)} \\
    \geq& \sum_{i=1}^M V_{1}^{(i)\star} \left( s_{1,t}^{(i)} \right) - M H \mcal{I}.
\end{align}
\qed 

\label{prf:lemma4}
\textbf{Lemma 4.} \textit{For any episode $t\in[T]$, level $h\in[H]$ and any Q-value function at next level $\{Q_{h+1}^{(i)}\}_{i=1}^M \in \mcal{Q}_{h+1}$, denote $\dot{f}_{h,t}$ as the best fit Q-value estimation induced by $Q_{h+1}^{(i)}$ minimizing Bellman error, we have}
\begin{align}
    \left\| \hat{f}_{h,t}\left[ Q_{h+1} \right] - \dot{f}_{h,t} \left[ Q_{h+1} \right] \right\|_{2,E_t}^2 \leq \beta_{h,t} \stackrel{\text { def }}{=} \left(B_{h,1} + \sqrt{MT} \mcal{I} + \sqrt{ B_{h,2}}\right)^2.
\end{align}
\textit{The $B_{h,1}$ and $B_{h,2}$ are from Lemma 6. Equivalently saying, this means that $\dot{f}_{h,t}$ is contained in set $\mcal{F}_{h,t}$ defined as}
\begin{align*}
    \mcal{F}_{h,t} \stackrel{\mathrm{def}}{=} \left\{ f\in\mcal{F}^{\otimes M}: \left\| f -\hat{f}_{h,t}\left[ Q_{h+1} \right] \right\|_{2,E_t}^2 \leq \beta_{h,t} \right\}.
\end{align*}

\textit{Proof.} By the empirical optimality of $\hat{f}_{h,t}$, we know
\begin{align}
    \sum_{i=1}^M \left\| \hat{f}_{h,t}^{(i)} (\bsyb{X}_{h,t}) - \bsyb{y}_{h,t}^{(i)} \right\|^2 \leq \sum_{i=1}^M \left\| \dot{f}_{h,t}^{(i)} (\bsyb{X}_{h,t}) - \bsyb{y}_{h,t}^{(i)} \right\|^2.
\end{align}
Here we abuse the notation and use $\hat{f}_{h,t}^{(i)} (\bsyb{X}_{h,t})$ to denote function $\hat{f}_{h,t}^{(i)}$'s output on all the state-action pair $\bsyb{X}_{h,t}$ in the first $t-1$ episodes at level $h$ for task $i$, also $\bsyb{y}_{h,t}^{(i)}$ is the corresponding target value label. This inequality implies that
\begin{align}
    &\sum_{i=1}^M \left\| \hat{f}_{h,t}^{(i)} (\bsyb{X}_{h,t}) - \dot{f}_{h,t}^{(i)} (\bsyb{X}_{h,t}) \right\|^2 \\
    \leq& 2\sum_{i=1}^M \left\langle \bsyb{\Delta}_{h,t}^{(i)}, \hat{f}_{h,t}^{(i)} (\bsyb{X}_{h,t}) - \dot{f}_{h,t}^{(i)} (\bsyb{X}_{h,t}) \right\rangle + 2\sum_{i=1}^M \left\langle \bsyb{z}_{h,t}^{(i)}, \hat{f}_{h,t}^{(i)} (\bsyb{X}_{h,t}) - \dot{f}_{h,t}^{(i)} (\bsyb{X}_{h,t}) \right\rangle 
\end{align}
where 
$$\bsyb{\Delta}_{h,t}^{(i)} \stackrel{\text{def}}{=} \left[ \Delta_{h,1}^{(i)}(Q_{h+1}^{(i)})(s_{h,1}^{(i)}, a_{h,2}^{(i)}) \quad  \Delta_{h,2}^{(i)}(Q_{h+1}^{(i)})(s_{h,2}^{(i)}, a_{h,2}^{(i)}) \quad \hdots \quad \Delta_{h,t-1}^{(i)}(Q_{h+1}^{(i)})(s_{h,t-1}^{(i)}, a_{h,t-1}^{(i)}) \right]$$ 
is the Bellman error for Q-value approximation, each $\Delta_{h,j}^{(i)}(Q_{h+1}^{(i)})(s_{h,j}^{(i)}, a_{h,j}^{(i)})$is defined in (30). And 
$$\bsyb{z}_{h,t}^{(i)} \stackrel{\text{def}}{=} \left[ z_{h,1}^{(i)}(Q_{h+1}^{(i)})(s_{h,1}^{(i)}, a_{h,2}^{(i)})\quad \hdots\quad z_{h,t-1}^{(i)}(Q_{h+1}^{(i)})(s_{h,t-1}^{(i)}, a_{h,t-1}^{(i)})  \right] $$ 
where $ z_{h,j}^{(i)} \left( Q_{h+1}^{(i)} \right) \left(s_{h,j}^{(i)}, a_{h,j}^{(i)} \right) \stackrel{\text{def}}{=} R\left( s_{h,j}^{(i)}, a_{h,j}^{(i)} \right) + \max_{a\in\mcal{A}} Q_{h+1}^{(i)} \left( s_{h+1,j}^{(i)}, a \right) - \mcal{T}_{h}^{(i)} \left( Q_{h+1}^{(i)} \right) \left( s_{h,j}^{(i)}, a_{h,j}^{(i)} \right) $ is the finite sampling noise. 

Next, we are going to bound the two terms in (58). For the first term, we have
\begin{align}
    & \sum_{i=1}^M \left\langle \bsyb{\Delta}_{h,t}^{(i)}, \hat{f}_{h,t}^{(i)} (\bsyb{X}_{h,t}) - \dot{f}_{h,t}^{(i)} (\bsyb{X}_{h,t}) \right\rangle \\
    \leq& \sum_{i=1}^M \left\| \bsyb{\Delta}_{h,t}^{(i)} \right\| \cdot \left\| \hat{f}_{h,t}^{(i)} (\bsyb{X}_{h,t}) - \dot{f}_{h,t}^{(i)} (\bsyb{X}_{h,t})\right\| \\
    \leq& \sqrt{T} \mcal{I} \cdot \sum_{i=1}^M \left\| \hat{f}_{h,t}^{(i)} (\bsyb{X}_{h,t}) - \dot{f}_{h,t}^{(i)} (\bsyb{X}_{h,t})\right\| \\
    \leq& \sqrt{MT} \mcal{I} \cdot \left\| \hat{f}_{h,t}  - \dot{f}_{h,t} \right\|_{2,E_t} 
\end{align}
By lemma 6, when the failure case does not happen, we have
\begin{align}
    \sum_{i=1}^M \left\langle \bsyb{z}_{h,t}^{(i)}, \hat{f}_{h,t}^{(i)} (\bsyb{X}_{h,t}) - \dot{f}_{h,t}^{(i)} (\bsyb{X}_{h,t}) \right\rangle \leq B_{h,1} \cdot \left\| \hat{f}_{h,t}  - \dot{f}_{h,t} \right\|_{2,E_t} + B_{h,2}
\end{align}
where 
\begin{align}
    B_{h,1} =& \sqrt{2 M k + \log( \mcal{N}(\Phi, (kMT)^{-1}, \|\cdot\|_{\infty} ) / \delta)} + 1\\
    B_{h,2} =& 2 \sqrt{M T + \log(2MT^2 / \delta)}
\end{align}
Adding the bound for two terms and we get
\begin{align}
    &\left\| \hat{f}_{h,t} - \dot{f}_{h,t} \right\|_{2,E_t}^2 \leq (B_{h,1} + \sqrt{MT} \mcal{I}) \cdot \left\| \hat{f}_{h,t} - \dot{f}_{h,t} \right\|_{2,E_t} + B_{h,2} \\
    \Longrightarrow\quad& \left\| \hat{f}_{h,t} - \dot{f}_{h,t} \right\|_{2,E_t}^2 \leq \left(B_{h,1} + \sqrt{MT} \mcal{I} + \sqrt{ B_{h,2}}\right)^2 \stackrel{\mathrm{def}}{=} \beta_{h,t}
\end{align}
which completes the proof.
\qed \\

\textbf{Lemma 5.} \textit{If the failure event in lemma 6 does not happen, for any feasible solution $Q_{h}^{(i)} \left[\tilde{f}_h^{(i)} \right]$ in the definition of $\mcal{F}_{h,t}$, and any $h \in [H]$, $t \in [T]$, we have}
\begin{align}
    \sum_{i=1}^M  \left| \left(\tilde{Q}_{h,t}^{(i)} - \mcal{T}_{h}^{(i)} \tilde{Q}_{h+1,t}^{(i)} \right) \left( s_{h,t}^{(i)}, a_{h,t}^{(i)} \right) \right| \leq M \mcal{I} + 2 w_{\mcal{F}_{h,t}} \left( \bsyb{x}_{h,t} \right),
\end{align}
\textit{where $\bsyb{x}_{h,t} = \left[(s_{h,t}^{(1)}, a_{h,t}^{(1)}), \hdots, (s_{h,t}^{(M)}, a_{h,t}^{(M)}) \right]$ denotes the stacked input for all state-action pair at level $h$, episode $t$.}

\textit{Proof.} 
\begin{align}
    & \sum_{i=1}^M  \left| \left(\tilde{Q}_{h,t}^{(i)} - \mcal{T}_{h}^{(i)} \tilde{Q}_{h+1,t}^{(i)} \right)\left( s_{h,t}^{(i)}, a_{h,t}^{(i)} \right) \right| \\
    =& \sum_{i=1}^M \left| \tilde{Q}_{h,t}^{(i)}(s,a) - \dot{f}_{h}^{(i)}\left[ \tilde{Q}_{h+1}^{(i)}\right]\left( s_{h,t}^{(i)}, a_{h,t}^{(i)} \right) - \Delta_{h}^{(i)} \left( \tilde{Q}_{h+1}^{(i)} \right) \left( s_{h,t}^{(i)}, a_{h,t}^{(i)} \right) \right| \\
    \leq& M\mcal{I} + \sum_{i=1}^M \left| \tilde{f}_{h,t}^{(i)}\left( s_{h,t}^{(i)}, a_{h,t}^{(i)} \right) - \dot{f}_{h}^{(i)}\left[ \tilde{Q}_{h+1}^{(i)}\right]\left( s_{h,t}^{(i)}, a_{h,t}^{(i)} \right) \right| \\
    \leq& M\mcal{I} + \sum_{i=1}^M \left| \tilde{f}_{h,t}^{(i)}\left( s_{h,t}^{(i)}, a_{h,t}^{(i)} \right) - \hat{f}_{h}^{(i)}\left( s_{h,t}^{(i)}, a_{h,t}^{(i)} \right) \right| + \left| \hat{f}_{h}^{(i)}\left( s_{h,t}^{(i)}, a_{h,t}^{(i)} \right) - \dot{f}_{h}^{(i)}\left[ \tilde{Q}_{h+1}^{(i)}\right]\left( s_{h,t}^{(i)}, a_{h,t}^{(i)} \right) \right| 
\end{align}
According to our construction, we know that both $\tilde{f}_{h,t}^{(i)}$ and $\dot{f}_{h}^{(i)}$ are contained in $\mcal{F}_{h,t}$, therefore we have $\sum_{i=1}^M \left| \tilde{f}_{h,t}^{(i)}\left( s_{h,t}^{(i)}, a_{h,t}^{(i)} \right) - \hat{f}_{h}^{(i)}\left( s_{h,t}^{(i)}, a_{h,t}^{(i)} \right)\right| \leq w_{\mcal{F}_{h,t}} \left( \bsyb{x}_{h,t} \right)$ and $\sum_{i=1}^M \left| \dot{f}_{h,t}^{(i)} \left[ \tilde{Q}_{h+1}^{(i)} \right] \left( s_{h,t}^{(i)}, a_{h,t}^{(i)} \right) - \hat{f}_{h}^{(i)}\left( s_{h,t}^{(i)}, a_{h,t}^{(i)} \right) \right| \leq w_{\mcal{F}_{h,t}}\left( \bsyb{x}_{h,t} \right) $, where $\bsyb{x}_{h,t} = \left[(s_{h,t}^{(1)}, a_{h,t}^{(1)}), \hdots, (s_{h,t}^{(M)}, a_{h,t}^{(M)}) \right]$ denotes the stacked input for all state-action pair at level $h$, episode $t$. 

Summarizing all the inequalities and we know the whole lemma holds.
\qed \\

\textbf{Lemma 6.} (Probability bound for failure event) \textit{In this lemma we denote $\hat{f}_{h}^{(i)}\left[ Q_{h+1}^{(i)} \right]$ as $\hat{f}_{h}^{(i)}$ for the sake of simplicity (similar for $\dot{f}_{h}^{(i)}$). Define event $E_{h,t}$ as}
\begin{align}
    E_{h,t} \stackrel{\text{def}}{=} \mathbb{I}\left[ \exists \{Q_{h+1}^{(i)} \}_{i=1}^M \quad \sum_{i=1}^M \left\langle \bsyb{z}_{h,t}^{(i)}, \hat{f}_{h}^{(i)}(\bsyb{X}_{h,t}) - \dot{f}_{h}^{(i)}(\bsyb{X}_{h,t} ) \right\rangle > B_{h,1} \cdot \left\| \hat{f}_{h}^{(i)} - \dot{f}_{h}^{(i)} \right\|_{2,E_t} + B_{h,2} \right]
\end{align}
where $B_{h,1}$ and $B_{h,2}$ will be specified later. We have
\begin{align}
    \mathbb{P}\left( \bigcup_{t=1}^T \bigcup_{h=1}^H E_{h,t} \right) \leq \delta.
\end{align}
\textit{Proof.} Similar to lemma 1, we can find a $\alpha$-cover $\Phi_{\alpha}$ for $\Phi$ such that for any Q-value function $\left( Q_{h+1}^{(1)}[ \phi \circ \bsyb{\theta}_1], Q_{h+1}^{(2)}[ \phi \circ \bsyb{\theta}_2], \hdots, Q_{h+1}^{(M)}[ \phi \circ \bsyb{\theta}_M] \right)$, we can find $\bar{\phi} \in \Phi_{\alpha}$ and $\bar{\bsyb{\theta}}_i$ for $i\in[M]$ such that for any $(s,a)\in \mcal{S} \times \mcal{A}$ and any $ i\in[M]$
\begin{align}
    \left| Q_{h+1}^{(i)}(s,a) - \bar{\phi}(s,a)^{\top} \bar{\bsyb{\theta}}_i  \right| \leq \sqrt{k} \alpha.
\end{align}
Define $\bar{Q}_{h+1}^{(i)} = Q_{h+1}^{(i)} \left[\bar{\phi}\circ \bsyb{\theta}_i \right]$ and further let 
$$\bar{\bsyb{z}}_{h,t}^{(i)} \stackrel{\text{def}}{=} \left[ z_{h,1}^{(i)} \left( \bar{Q}_{h+1}^{(i)} \right) \left(  s_{h,1}^{(i)}, a_{h,1}^{(i)} \right) \quad \hdots \quad z_{h,t-1}^{(i)} \left( \bar{Q}_{h+1}^{(i)} \right) \left( s_{h,t-1}^{(i)}, a_{h,t-1}^{(i)} \right) \right] \in \mathbb{R}^{t-1} $$
then we have
\begin{align}
    & \sum_{i=1}^M \left\langle \bsyb{z}_{h,t}^{(i)}, \hat{f}_{h}^{(i)}(\bsyb{X}_{h,t}) - \dot{f}_{h}^{(i)}(\bsyb{X}_{h,t} ) \right\rangle \\
    =& \sum_{i=1}^M \left\langle \bar{\bsyb{z}}_{h,t}^{(i)}, \hat{f}_{h}^{(i)}(\bsyb{X}_{h,t}) - \dot{f}_{h}^{(i)}(\bsyb{X}_{h,t} ) \right\rangle \\
    +&\sum_{i=1}^M \left\langle \bsyb{z}_{h,t}^{(i)} - \bar{\bsyb{z}}_{h,t}^{(i)}, \hat{f}_{h}^{(i)}(\bsyb{X}_{h,t}) - \dot{f}_{h}^{(i)}(\bsyb{X}_{h,t} ) \right\rangle \\
\end{align}
Notice that for fixed $\bar{f}_{h}^{(i)}(\cdot, \cdot) = \phi(\cdot, \cdot)^{\top} \bar{\bsyb{\theta}}_{h+1}^{(i)}$, each $ z_{h,1}^{(i)} \left( \bar{Q}_{h+1}^{(i)} \right) \left(  s_{h,1}^{(i)}, a_{h,2}^{(i)} \right)$ is a zero-mean 1-sub-Gaussian random variable conditioned on past history. Therefore we can treat it as $\eta_{t,i} = z_{h,t}^{(i)}$ in Lemma 1 and get
\begin{align}
  & \sum_{i=1}^M \left\langle \bar{\bsyb{z}}_{h,t}^{(i)}, \hat{f}_{h}^{(i)}(\bsyb{X}_{h,t}) - \dot{f}_{h}^{(i)}(\bsyb{X}_{h,t} ) \right\rangle  \\ 
  \leq& \sqrt{2 M k + \log(1 / \delta_1)} \left\| \hat{f}_{h,t} - \dot{f}_{h,t} \right\|_{2,E_t} + 2\alpha \sqrt{M t k (M t + \log(2Mt^2 / \delta_2))}.
\end{align}
Setting $\delta_1 = \frac{\delta}{2|\Phi^{\alpha}|}, \delta_2 = \delta/2$ and get
\begin{align}
  & \sum_{i=1}^M \left\langle \bar{\bsyb{z}}_{h,t}^{(i)}, \hat{f}_{h}^{(i)}(\bsyb{X}_{h,t}) - \dot{f}_{h}^{(i)}(\bsyb{X}_{h,t} ) \right\rangle  \\ 
  \leq& \sqrt{2 M k + \log( \mcal{N}(\Phi, \alpha, \|\cdot\|_{\infty} ) / \delta)} \cdot \left\| \hat{f}_{h,t} - \dot{f}_{h,t} \right\|_{2,E_t} + 2\alpha \sqrt{M T k (M T + \log(2MT^2 / \delta))}.
\end{align}
By union bound, we know it holds for any $\bar{f}_{h}$ with probability at least $1-|\Phi^{\alpha}| \delta_1=1-\delta$. Also, from $ \left| Q_{h+1}^{(i)}(s,a) - \bar{\phi}(s,a)^{\top} \bar{\bsyb{\theta}}_i  \right| \leq \sqrt{k} \alpha'$ we know that
\begin{align}
 & \left| z_{h,j}^{(i)} \left( Q_{h+1}^{(i)} \right) \left(  s_{h,j}^{(i)}, a_{h,j}^{(i)} \right) - z_{h,j}^{(i)} \left( \bar{Q}_{h+1}^{(i)} \right) \left(  s_{h,j}^{(i)}, a_{h,j}^{(i)} \right) \right| \\    
 =& \left| \max_{a\in\mcal{A}} Q_{h+1}^{(i)} \left( s_{h+1,j}^{(i)}, a \right) - \mcal{T}_{h}^{(i)} \left( Q_{h+1}^{(i)} \right) \left( s_{h,j}^{(i)}, a_{h,j}^{(i)} \right) - \max_{a\in\mcal{A}} \bar{Q}_{h+1}^{(i)} \left( s_{h+1,j}^{(i)}, a \right) + \mcal{T}_{h}^{(i)} \left( \bar{Q}_{h+1}^{(i)} \right) \left( s_{h,j}^{(i)}, a_{h,j}^{(i)} \right) \right| \\
 \leq& \max_{a\in\mcal{A}} \left|  Q_{h+1}^{(i)} \left( s_{h+1,j}^{(i)}, a \right) -  \bar{Q}_{h+1}^{(i)} \left( s_{h+1,j}^{(i)}, a \right) \right| + \left| \mcal{T}_{h}^{(i)} \left( \bar{Q}_{h+1}^{(i)} -  Q_{h+1}^{(i)}  \right) \left( s_{h,j}^{(i)}, a_{h,j}^{(i)} \right) \right| \\
 \leq& 2\sqrt{k} \alpha'
\end{align}

hence we have
\begin{align}
    &\sum_{i=1}^M \left\langle \bsyb{z}_{h,t}^{(i)} - \bar{\bsyb{z}}_{h,t}^{(i)}, \hat{f}_{h}^{(i)}(\bsyb{X}_{h,t}) - \dot{f}_{h}^{(i)}(\bsyb{X}_{h,t} ) \right\rangle \\
    \leq& \sum_{i=1}^M \left\| \bsyb{z}_{h,t}^{(i)} - \bar{\bsyb{z}}_{h,t}^{(i)} \right\| \cdot \left\| \hat{f}_{h}^{(i)}(\bsyb{X}_{h,t}) - \dot{f}_{h}^{(i)}(\bsyb{X}_{h,t} ) \right\| \\
    \leq& 2 \alpha' \sqrt{M T k} \cdot \left\|\hat{f}_{h,t}- \dot{f}_{h,t} \right\|_{2,E_t}
\end{align}
holds for arbitrary $\{Q_{h+1}^{(i)}\}$ at any level $h\in[H], t \in [T]$.

Adding (83) and (90), we finally finish the proof by setting $\alpha=\alpha'= \frac{1}{M T k}$
\begin{align}
    B_{h,1} =& \sqrt{2 M k + \log( \mcal{N}(\Phi, (kMT)^{-1}, \|\cdot\|_{\infty} ) / \delta)} + 1\\
    B_{h,2} =& 2 \sqrt{M T + \log(2MT^2 / \delta)}
\end{align}

\qed \\

\section{Experiment Dissection and Discussion}
In this section, we will take a closer view of the learning procedure and analyze the functionality of the UCB term in our algorithm. Usually, a reasonable UCB term should embrace several properties. \textit{(i)} It should let confidence set $\mcal{F}_t$ contain the real parameter with high probability. \textit{(ii)} It should shrink at a reasonable speed to achieve low regret. 

\begin{figure*}[ht]
\centering
\label{fig:errbonus}
\subfigure[]{
\includegraphics[width=0.48\linewidth]{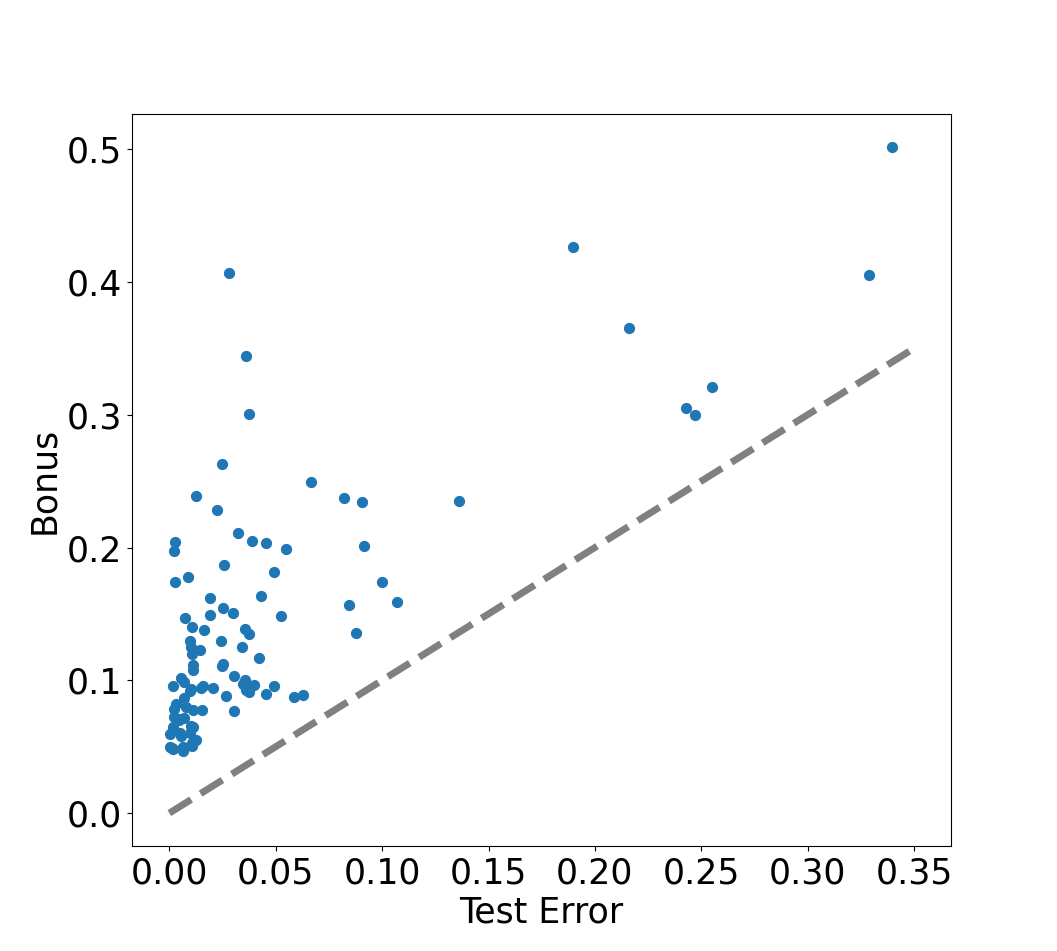}
}
\label{fig:shrink}
\subfigure[]{
\includegraphics[width=0.48\linewidth]{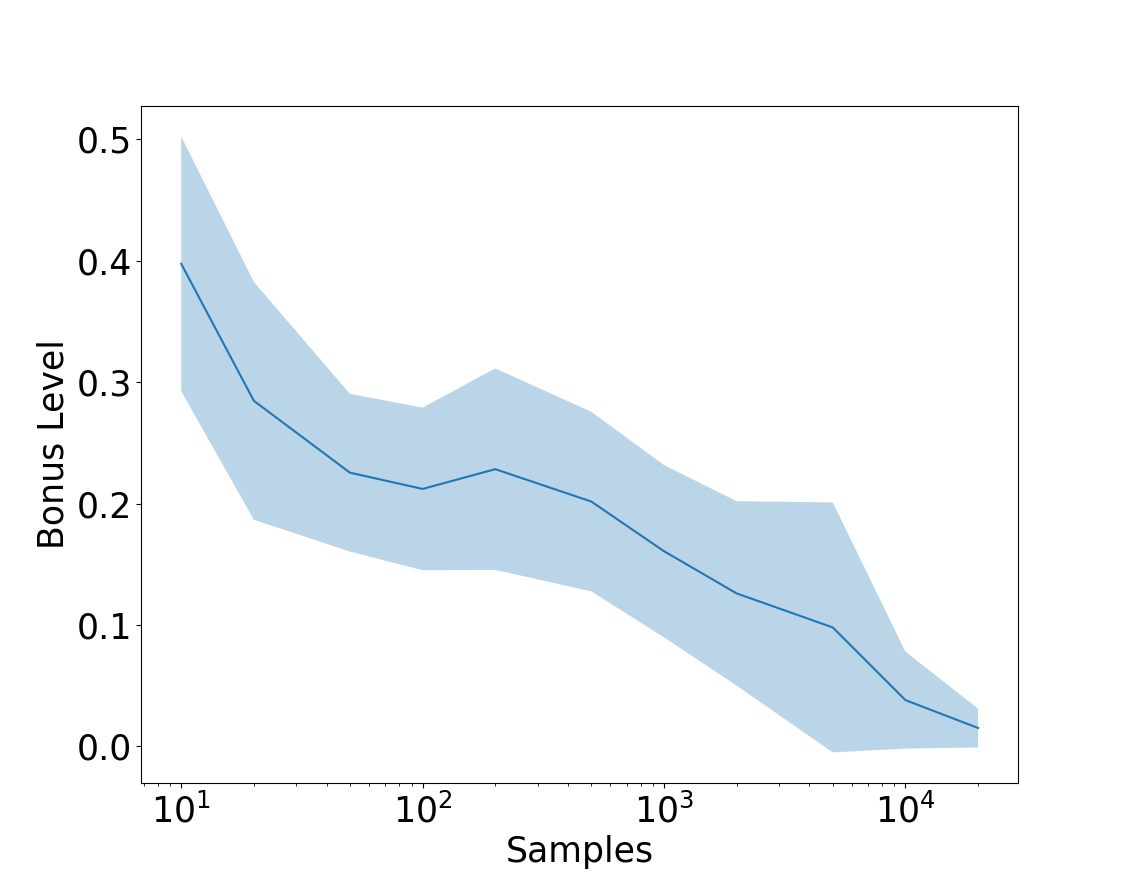}
}
\caption{(a) The relationship between unknown data's prediction error and the bonus it gets from finetuning. The grey line is $y=x$. (b) The average bonus level of 100 test images with respect to the number of samples in training set, the shaded area is the interval for $\pm 1$ standard deviation.}
\end{figure*}

To check (i), we choose the model $\hat{f}_{t}$ at step $t=200$ which is trained on insufficient data with only 2000 samples. We then sample $100$ images from test set as unknown inputs $\mcal{D}=\{(\bsyb{x}_i, y_i)\}_{i=1}^{100}$, where $\bsyb{x}_i$ is the digit image and $y_i$ is the corresponding target value. We inspect the relationship between the original prediction error $|\hat{f}_t(\bsyb{x}_i)-y_i|$ and the added bonus $b_i = \bar{f}_t(\bsyb{x}_i)-\hat{f}_t(\bsyb{x}_i)$ via finetuning on each input $\bsyb{x}_i \in \mcal{D}$. The result is presented as scatter dots in \hyperref[fig:errbonus]{Figure 2(a)}. We can clearly see that almost all the points lie above the line $y=x$, meaning that $b_i=\bar{f}_t(\bsyb{x}_i)-\hat{f}_t(\bsyb{x}_i)\geq |\hat{f}_t(\bsyb{x}_i)-y_i| \geq y_i- \hat{f}_t(\bsyb{x}_i) $ for any $i\in[100]$, which further indicates that $\bar{f}_t(\bsyb{x}_i) \geq y_i$. This validates that we can always find some $\bar{f}\in\mcal{F}_t$ to give an optimistic estimation of the value for almost every $\bsyb{x}$. Moreover, we can observe an apparent correlated pattern between the test error and bonus, which implies that our algorithm will give larger bonus for the data point whose prediction is not reliable, and only give relatively small bonus for the data that it is confident with.

We also check (ii) by plotting the average bonus level (closely related to the width of confidence set) against the number of samples the algorithm has been trained on. We gradually increase the number of samples from $10$ to $20000$ and fix a set of test images $\mcal{D}$ as before to see how the average bonus level changes when the training set size increases. The result is shown in \hyperref[fig:shrink]{Figure 2(b)}. Previous work \cite{dong2021provable} proves that the eluder dimension of neural networks can be exponentially large in the worst case, which means that it can give almost arbitrary output value even when it is constrained to give a precisely accurate prediction for a large number of samples in the training set. In that case, the average bonus level should have remained constant regardless of the size of the training set. However, our experiment shows that the average bonus drops when the number of training samples increases. We conjecture that it is because in reality, when the input data are restricted to regular images with clear semantics, and the optimization procedure of the model is conducted via gradient-based methods in a very close neighborhood, the arbitrariness of the neural network's output is substantially reduced. 

Restricting the model's training loss in the training set effectively limits the bonus obtained from the finetune procedure, which realizes the desired fast-shrinking property from our functional confidence set. Such a phenomenon sheds light on the unknown property of neural network's generalization capability and interpolation plasticity. We leave explaining the underlying mechanism as future work. 

\subsection{Visualize the Learned Representation}
\label{apdx:visualize}
A natural and interesting question is what representation does our CNN backbone actually learn. To investigate this problem and visualize the learned representation, we measure the information of different digits within the learned representation. Interestingly, we find that our model indeed learns an indicative representation for classification problem via multitask value regression training. 

The basic measurement for the quality of representation is evaluated with the kernel function $\kappa(\bsyb{x}_i, \bsyb{x}_j) = \left\langle \phi(\bsyb{x}_i), \phi(\bsyb{x}_j) \right\rangle$ and see whether it has a strong diagonal. We take the checkpoint of neural network model at final step (around 600 with more than 6000 samples), and treat the module before the final linear layer as $\phi(\cdot)$. Denote the MNIST test set as $\mcal{D}=\{\mcal{D}_i\}_{i=0}^9$ where $\mcal{D}_i$ is the images of digit $i$. Define the correlation between digit $i$ and $j$ under representation $\phi$ as 
\begin{align}
    C(i,j) = \frac{1}{|\mcal{D}_i|\times|\mcal{D}_j|}\sum_{\bsyb{x}_s \in \mcal{D}_i} \sum_{\bsyb{x}_t \in \mcal{D}_j} \left\langle \phi(\bsyb{x}_s), \phi(\bsyb{x}_t) \right\rangle
\end{align}
To accelerate the evaluation, notice that we can preprocess an ``template vector'' $\bsyb{T}_i$ for each digit $i$ as
\begin{align}
    \bsyb{T}_i = \frac{1}{|\mcal{D}_i|} \sum_{\bsyb{x}\in\mcal{D}_i} \phi(\bsyb{x})
\end{align}
so that the correlation can be computed through 
\begin{align}
    C(i,j) =& \frac{1}{|\mcal{D}_i|\times|\mcal{D}_j|}\sum_{\bsyb{x}_s \in \mcal{D}_i} \sum_{\bsyb{x}_t \in \mcal{D}_j} \left\langle \phi(\bsyb{x}_s), \phi(\bsyb{x}_t) \right\rangle \\
    =& \frac{1}{|\mcal{D}_j|} \sum_{\bsyb{x}_t \in \mcal{D}_j} \left( \frac{1}{|\mcal{D}_i|} \sum_{\bsyb{x}_s \in \mcal{D}_i} \left\langle \phi(\bsyb{x}_s), \phi(\bsyb{x}_t) \right\rangle  \right)\\
    =& \frac{1}{|\mcal{D}_j|}  \sum_{\bsyb{x}_t \in \mcal{D}_j} \left\langle \frac{1}{|\mcal{D}_i|}\sum_{\bsyb{x}_s \in \mcal{D}_i}\phi(\bsyb{x}_s), \phi(\bsyb{x}_t) \right\rangle  \\
    =& \frac{1}{|\mcal{D}_j|}  \sum_{\bsyb{x}_t \in \mcal{D}_j}\left\langle \bsyb{T}_i, \phi(\bsyb{x}_t) \right\rangle \\
    =& \left\langle \bsyb{T}_i, \bsyb{T}_j\right\rangle
\end{align}

We plot this 10x10 correlation map for single task training and multitask training with $M=10$. Notice that the single task reward mapping function is $\sigma(i)=i/10$, and to assure the different tasks in multitask training are heterogeneous, we manually set that the best digit for each task are distinct. 

The result is in figure 3. We can see that since single task only needs to recognize the large value digit, namely 9, 8 or 7, its representation function is not informative for distinguishing digits. And interestingly, the multitask trained network's representation demonstrates a very strong diagonal, indicating that the representation vector is very specific to the digit's image, although the training process has no explicit definition for the classification task but a regression problem instead. Actually, we found a simple linear layer append to this representation can achieve over 95$\%$ accuracy on MNIST test set.
\begin{figure*}[ht]
\centering
\label{fig:rep1}
\subfigure[Single task]{
\includegraphics[width=0.43\linewidth]{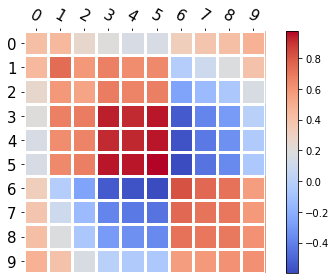}
}
\label{fig:rep2}
\subfigure[Multitask $M=10$]{
\includegraphics[width=0.43\linewidth]{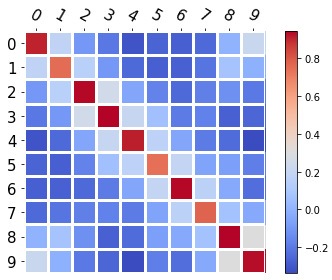}
}
\caption{The kernel function for the representation learned by single task and 10-tasks multitask. It is clear that multitask representation learning obtains a more comprehensive and interpretable pattern for the MNIST images.}
\end{figure*}

\end{document}